\documentclass{article}

\usepackage{microtype}
\usepackage{graphicx}
\usepackage{subfigure}
\usepackage{wrapfig}
\usepackage{booktabs} 

\usepackage{hyperref}

\usepackage[accepted]{icml2024}

\usepackage{amsmath}
\usepackage{amssymb}
\usepackage{mathtools}
\usepackage{amsthm}
\usepackage{multirow}
\usepackage{graphicx}

\usepackage[capitalize,noabbrev]{cleveref}

\theoremstyle{plain}
\newtheorem{theorem}{Theorem}[section]

\newtheorem{lemma}[theorem]{Lemma}

\theoremstyle{definition}

\theoremstyle{remark}

\usepackage[textsize=tiny]{todonotes}

\usepackage{amsmath,amsfonts,bm}

\def\eqref#1{equation~\ref{#1}}

\def\1{\bm{1}}

\def\eps{{\epsilon}}

\def\mA{{\mathbf{A}}}
\def\mB{{\mathbf{B}}}

\def\mK{{\mathbf{K}}}

\def\mM{{\mathbf{M}}}

\def\mQ{{\mathbf{Q}}}

\def\mS{{\mathbf{S}}}

\def\mW{{\mathbf{W}}}
\def\mX{{\mathbf{X}}}

\def\va{{\bm{a}}}

\def\ve{{\bm{e}}}
\def\vf{{\bm{f}}}
\def\vg{{\bm{g}}}

\def\vk{{\bm{k}}}

\def\vm{{\bm{m}}}

\def\vo{{\bm{o}}}

\def\vq{{\bm{q}}}

\def\vu{{\bm{u}}}

\def\vx{{\bm{x}}}
\def\vy{{\bm{y}}}

\DeclareMathAlphabet{\mathsfit}{\encodingdefault}{\sfdefault}{m}{sl}
\SetMathAlphabet{\mathsfit}{bold}{\encodingdefault}{\sfdefault}{bx}{n}
\newcommand{\tens}[1]{\bm{\mathsfit{#1}}}

\def\tM{{\tens{M}}}

\def\gN{{\mathcal{N}}}

\def\gP{{\mathcal{P}}}
\def\gQ{{\mathcal{Q}}}

\newcommand{\R}{\mathbb{R}}

\def\R{{\boldsymbol R}}

\icmltitlerunning{Bilevel Positional Encoding for Better Length Extrapolation}

\begin{document}

\twocolumn[
\icmltitle{Two Stones Hit One Bird: Bilevel Positional Encoding \\for Better Length Extrapolation}



\icmlsetsymbol{equal}{*}

\begin{icmlauthorlist}
\icmlauthor{Zhenyu He}{equal,pku_ai}
\icmlauthor{Guhao Feng}{equal,pku_cs}
\icmlauthor{Shengjie Luo}{equal,pku_ai}
\icmlauthor{Kai Yang}{pku_cs}\\
\icmlauthor{Liwei Wang}{pku_ai,pku_ml}
\icmlauthor{Jingjing Xu}{bytedance}
\icmlauthor{Zhi Zhang}{bytedance}
\icmlauthor{Hongxia Yang}{bytedance}
\icmlauthor{Di He}{pku_ai}
\\ \vspace{4pt}\emph{Code}: \url{https://github.com/zhenyuhe00/BiPE}
\vspace{-12pt}

\end{icmlauthorlist}

\icmlaffiliation{pku_ai}{National Key Laboratory of General Artificial Intelligence, School of Intelligence Science and Technology, Peking University}

\icmlaffiliation{pku_cs}{School of EECS, Peking University}

\icmlaffiliation{pku_ml}{Center for Machine Learning Research, Peking University}

\icmlaffiliation{bytedance}{ByteDance Inc}

\icmlcorrespondingauthor{Di He}{dihe@pku.edu.cn}

\icmlkeywords{Length Extrapolation, Positional Encoding, Transformer}

\vskip 0.3in
]



\printAffiliationsAndNotice{\icmlEqualContribution} 

\begin{abstract}
In this work, we leverage the intrinsic segmentation of language sequences and design a new positional encoding method called Bilevel Positional Encoding (BiPE). 
For each position, our BiPE blends an intra-segment encoding and an inter-segment encoding. The intra-segment encoding identifies the locations within a segment and helps the model capture the semantic information therein via absolute positional encoding. The inter-segment encoding specifies the segment index, models the relationships between segments, and aims to improve extrapolation capabilities via relative positional encoding. Theoretical analysis shows this disentanglement of positional information makes learning more effective. The empirical results also show that our BiPE has superior length extrapolation capabilities across a wide range of tasks in diverse text modalities.
\end{abstract}

\vspace{-20pt}
\section{Introduction}

In many scenarios, text can be effectively decomposed into modular segments, each expressing a self-contained unit of thought~\cite{halliday2013halliday}. In natural languages, documents are typically composed of sentences. Each sentence describes a distinct idea or argument. In programming languages, code is organized into lines or function classes that define coherent operation or functionality. In mathematics, proofs unfold through a series of deductive steps, each representing a logical progression from its predecessors to the final answer.

\begin{figure*}[t]
    \includegraphics[width=0.995\linewidth]{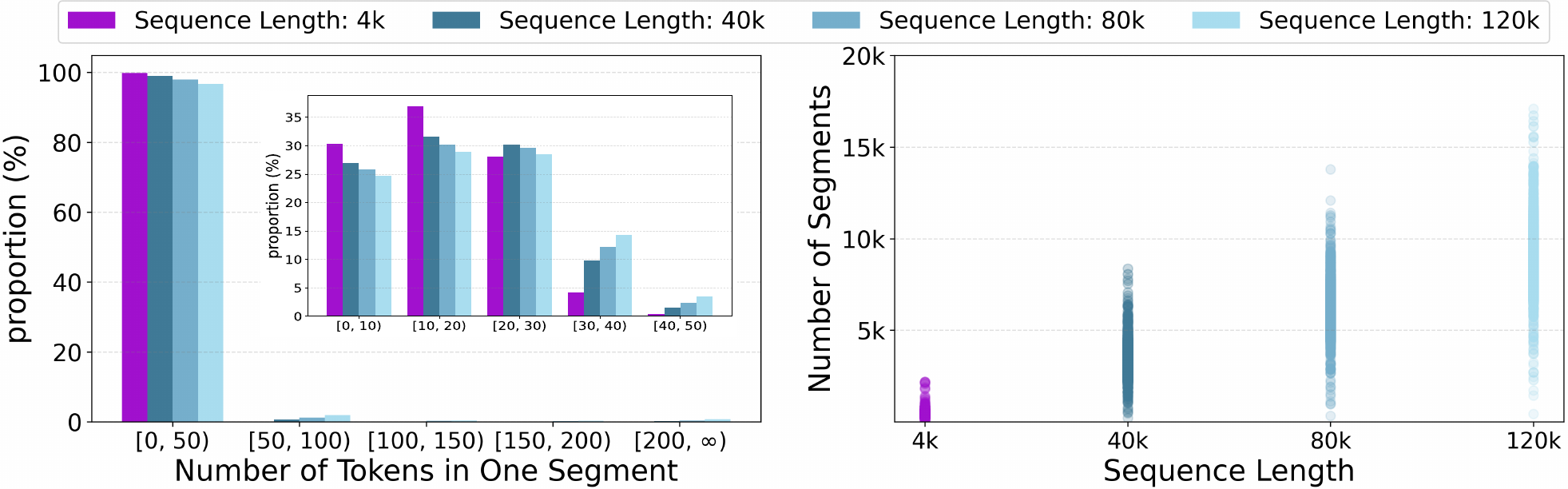}
    \vspace{-5pt}
    \caption{\textbf{Left:} The distribution of the token number in one segment with different sequence lengths. \textbf{Right:} The distribution of the number of segments with different sequence lengths.  We use the tokenizer of Llama 2~\cite{touvron2023llama} for tokenization on PG-19~\cite{Rae2020Compressive}. Full stop``\texttt{.}'' and newline ``\texttt{\textbackslash n}'' are used for segmentation. It can be seen that even when the sequence length is around 120k, the token number in most sentences is less than 50, while the number of sentences grows up to 10k.
    }
    \vspace{-8pt}
    \label{fig:token_distribution}
\end{figure*}

The lengths of different text sequences may vary significantly~\cite{alibi}. What is intriguing is that empirically, we observed that for sequences with different lengths, the distribution of the token number in each modular segment usually has bounded support and tends to be approximately similar. In Figure~\ref{fig:token_distribution}, we utilized the widely used PG-19 text corpus~\cite{Rae2020Compressive} for visualization (Please refer to Appendix~\ref{app_exp:add_distribution} for more results). It is evident that the token number distribution in each segment (i.e., sentence) remains remarkably consistent, regardless of the total sequence length. In contrast, the number of sentences linearly increases as the sequence length grows.

\looseness=-1Given the above observations, we argue that a popular research direction in language modeling, known as the \emph{length extrapolation} problem~\cite{alibi,anil2022exploring,chi2022kerple,chi2023dissecting,chowdhury2023monotonic,chen2023extending}, should be better positioned as a \emph{number-of-segment extrapolation} problem. In the literature, previous studies in this direction have either developed better positional encodings that can handle longer sequence~\cite{raffel2020exploring,rope,randompos,kazemnejad2023impact,chen2023extending,li2023functional,peng2023yarn,zhu2023pose,chen2023clex,liu2023scaling} or proposed specific inductive biases associated with the attention patterns in language models~\cite{ratner2023parallel,han2023lminfinite,xiao2023efficient}, or both~\cite{alibi,chi2022kerple,chi2023dissecting,sun2022length}. However, none of these methods adequately consider or utilize the intrinsic segmentation of language data, nor do they specifically address the extrapolation issue in terms of the number of segments.

In this paper, we introduce \textbf{BiPE} (\underline{\textbf{Bi}}level \underline{\textbf{P}}ositional \underline{\textbf{E}}ncoding), a simple yet effective positional encoding scheme for improving length extrapolation. Different from all existing length extrapolation approaches, BiPE employs two distinct encodings for each position: an intra-segment encoding and an inter-segment encoding. The intra-segment encoding identifies the location of the token within its segment. As a complement, the inter-segment encoding specifies the segment to which it belongs. Using natural language as an illustration, different words within the same sentence share the same inter-segment positional encoding but possess different intra-segment encodings. Conversely, words in different sentences but occupying the same inter-segment position (e.g., the first token in different sentences) share the same intra-segment encoding while having distinct inter-segment encodings. See Figure~\ref{fig:1_position} for an illustration.

BiPE disentangles the position modeling, offering greater flexibility in addressing the length extrapolation problem. At the intra-segment level, the intra-segment encoding specifies positions within the segment, helping the model capture the semantic information contained therein. Given that the number of tokens within a segment is usually bounded, we discovered that utilizing the original absolute positional encodings (APE,~\citealp{vaswani2017attention}) is already sufficient at this level. The inter-segment encoding targets to capture the relationships between segments and exhibits certain extrapolation capabilities. Therefore, we employ relative positional encodings (RPE,~\citealp{rope,alibi}). In this way, inductive biases in the two levels focus on different aspects of positional information and can be appropriately incorporated into model architectures, leading to a better learning process. We further give a theoretical justification of BiPE, which suggests that the proposed positional encoding scheme can make the Transformer model more parameter-efficient under some conditions.

\looseness=-1Extensive experiments are conducted to demonstrate the empirical effectiveness of BiPE. First, we empirically verify the expressiveness of BiPE in mathematical reasoning tasks~\cite{wei2022chain,feng2023towards}, which well aligns with our theoretical results. Second, for the length extrapolation problem, we evaluate BiPE and strong baselines across diverse tasks covering language modeling [PG-19~\cite{Rae2020Compressive}, ArXiv and Github~\cite{gao2020pile}] and long context benchmark [SCROLLS~\cite{shaham2022scrolls}]. Finally, we conduct experiments on datasets of normal-length sentences. We also conduct ablation studies to verify the effectiveness of each module in BiPE. Our empirical results show the superior performance of BiPE on most problems. 
\section{Related Work}
This work focuses on the \emph{length extrapolation} problem in language modeling, i.e., can a language model that is trained on sequences with maximum length $L_{\text{train}}$ still perform well when being tested on sequences with length $L_{\text{test}}>L_{\text{train}}$?~\citep{alibi}. Here we provide a literature review of existing approaches related to this problem.

\subsection{Improved Positional Encodings for Length Extrapolation}
The original Transformer model~\cite{vaswani2017attention} encodes position information via Absolute Positional Encoding (APE), where each position is equipped with a (learnable or fixed sinusoidal) real-valued embedding. But neither the learnable nor the fixed sinusoidal embedding can generalize well to longer sequences.

Different from APE that assigns an embedding for each position $i$, \citet{shaw2018self} introduced Relative Positional Encoding (RPE) which encodes the relative distance $i-j$ for each position pair $(i,j)$. Most methods incorporate RPE as an additive term in the attention module~\citep{raffel2020exploring, alibi, chi2022kerple,chi2023dissecting}. These methods can mitigate the length extrapolation problem to some extent but still have several limitations. For example, \citet{raffel2020exploring} uses the same attention bias for all query-key pairs with a relative distance larger than $K$, which limits its ability to distinguish different positions in long sequences. 

One of the most popularly used relative positional encoding in recent large language models is Rotary Position Encoding (RoPE)~\citep{rope, chowdhery2022palm,touvron2023llama}. RoPE rotates the query and key vectors with an angle proportional to their absolute positions before the attention, which results in the attention being a function of the relative distance between tokens. 

While encoder-only Transformers (e.g., BERT \citep{devlin-etal-2019-bert}) are permutation equivariant without positional encoding, \citet{haviv-etal-2022-transformer} show that decoder-only Transformers with causal attention masks can learn positional information even without any explicit positional encoding. Recently, \citet{kazemnejad2023impact} discovered that the no positional encoding (NoPE) model also can handle longer sequences to some extent on small-scale synthetic tasks, but there is no strongly positive evidence on large-scale settings.

\vspace{-4pt}
\subsection{Improved Algorithms for Length Extrapolation}
To help positional encodings handle longer sequences, \citet{ruoss-etal-2023-randomized} recently proposed a way to randomly select a subset of positions from a much larger range than those observed during training. The positional information of longer sequences can thus be simulated. \citet{zhu2023pose} proposed a similar idea called positional skip-wise fine-tuning (PoSE), which requires additional efforts for fine-tuning large-scale models. 

Relative positional encoding, especially RoPE, can capture the relative positional information well, but its length extrapolation capability is not satisfactory yet. Due to this, one line of works introduces priors biased toward local window attention via additive RPEs~\citep{alibi,chi2022kerple,chi2023dissecting,sun2022length} or hard constraints~\citep{ratner2023parallel,xiao2023efficient,han2023lminfinite} to boost length extrapolation capabilities. Another line of works tailored to RoPE called positional embedding scaling~\citep{chen2023extending,peng2023yarn,roziere2023code,chen2023clex,liu2023scaling} adjusts the range of either the position index or the frequency basis in RoPE, achieving promising extrapolation performance. Recently, a concurrent work~\citep{jin2024llm} proposed a bilevel attention mechanism for better length extrapolation of RoPE-based language models. It keeps the exact attention computation within a pre-defined neighbor range and uses the floor operation to group and map unseen large relative positions. 

BiPE aims to develop a new positional encoding scheme, which is orthogonal to all the methods above. All these advancements can be seamlessly combined with BiPE for better length extrapolation. Our experiments (Section \ref{sec:exp}) on several representative algorithms provide strong evidence supporting the compatibility of BiPE.
\begin{figure*}[t]
\centering
\includegraphics[width=0.995\linewidth]{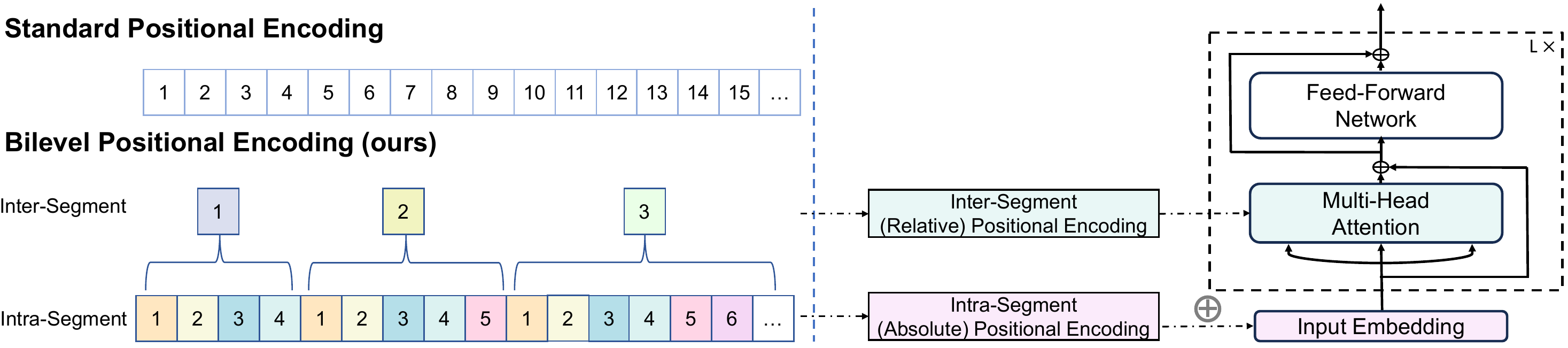}
\caption{
    \textbf{Left:} The schematic comparison of Standard Positional Encoding (top) and our proposed Bilevel Positional Encoding (BiPE, bottom). BiPE differentiates positions using both intra-segment and inter-segment encodings. \textbf{Right:} Absolute positional encoding is used as Intra-Segment Encoding added to the input embedding and relative positional encoding (e.g., RoPE and ALiBi) is used as Inter-Segment Encoding in the Transformer attention module.
}
\vspace{-4pt}
\label{fig:1_position}
\vspace{-10pt}
\end{figure*}

\vspace{-4pt}
\section{Method}
In this section, we introduce \textbf{BiPE} (\underline{\textbf{Bi}}level \underline{\textbf{P}}ositional \underline{\textbf{E}}ncoding), a new positional encoding scheme for better length extrapolation capabilities. First, we formally describe the modular segments of text sequence in Section \ref{sec:seq-rep}. Based on this segment representation, we thoroughly illustrate the derivation of our BiPE method in Section \ref{sec:BiPE}. In Section \ref{sec:theory}, we conduct a theoretical study on the expressiveness of our BiPE to further demonstrate its soundness.

\vspace{-5pt}
\subsection{Modular Segments of Text Sequence}\label{sec:seq-rep}
\vspace{-3pt}
\looseness=-1Formally, we use $\mathbf{S} = [w_1, \dots, w_l, \dots, w_L]$ to denote the input text sequence, i.e., an ordered collection of text tokens where $w_l$ denotes the $l$-th token and $L$ denotes the total sequence length. As previously introduced, text sequences can be decomposed into a series of non-overlapping modular segments, i.e., $\mathbf{S} = S_1 \oplus S_2 \oplus \dots S_n \oplus \dots \oplus S_N$ where $S_n$ is the $n$-th segment, $N$ is the total number of segments, and $\oplus$ is the concatenation operation. Each segment $S_{n}$ is defined by $S_{n} = [w_{a_{n}}, w_{a_{n}+1}, \dots, w_{b_{n}}]$. Here, $a_{n}$ and $b_{n}$ denote the starting and ending indices. The segmentation strategy can simply use symbol detection (e.g., newline and full stop). 

\vspace{-3pt}
\subsection{Bilevel Positional Encoding}\label{sec:BiPE}
\vspace{-3pt}
As stated in the introduction, in many practical scenarios, the number of tokens in each segment $S_n$ follows a similar distribution regardless of the value $L$, and the sequence length $L$ has a major impact on the segment number $N$. Consequently, we believe modeling the length extrapolation for $N$ is a more effective approach than that for actual length $L$. To this end, we propose BiPE, a novel bilevel positional encoding that blends two distinct encoding schemes at each position for better length extrapolation: an intra-segment encoding and an inter-segment encoding. (See Figure~\ref{fig:1_position}).

\textbf{Intra-Segment Encoding}. In a text sequence, each modular segment describes an independent statement, and the intra-segment positional encoding serves as an anchor to identify the location of each token in the segment for capturing semantic information therein. Formally, within each segment $S_{n}=[w_{a_{n}}, w_{a_{n}+1}, $ $ \dots,w_{b_{n}}]$, we encode the (local) position $i$ for token $w_{a_{n}+i}$, where $1\leq i\leq b_n-a_n+1$. Note that the number of tokens within a segment is usually bounded, i.e., there are few sentences that are extremely long. We find using the original absolute positional encoding~\citep{vaswani2017attention} is enough.
For each token $w_{a_{n}+i}$ in $S_{n}$, we assign a real-valued embedding $\ve_i$ to it, which will be added to the input token embedding. $\ve_i$ is shared among tokens at the same local position $i$ in different $S_n$ .

\textbf{Inter-Segment Encoding}. Though the intra-segment encoding can provide the location of tokens within each segment, the locations across segments are mixed, and the contextual relationships between segments are not captured. As a complement, we use the inter-segment positional encoding to specify the segment to which each token belongs. Keeping in mind that this encoding will play another role in handling longer sequences that are unseen during training, we employ relative positional encodings~\citep{shaw2018self,raffel2020exploring,rope,alibi}. Different from previous RPEs that are defined using the distance between token indexes,  the inter-segment encoding is defined using the distance between segment indexes.

\textbf{Instantiation}. BiPE uses absolute positional information for the intra-segment encoding and can leverage any RPE approaches for the inter-segment encoding. In this work, we instantiate two BiPE variants, BiPE-RoPE and BiPE-ALiBi.

BiPE-RoPE leverages RoPE~\cite{rope} as the inter-segment encoding. For a pair of tokens $(w_{l_1}, w_{l_2})$ which are in the $n$-th segment and the $m$-th segment respectively, two rotation matrices $\R_{\Theta, n}$ and $\R_{\Theta, m}$ are assigned, where $\Theta$ denotes the pre-defined parameters of the rotation matrix~\citep{rope}. Given query-key pair $q_{l_1},k_{l_2}\in\mathbb{R}^d$, the attention score is computed by $\frac{q_{l_1}\R_{\Theta, n}(k_{l_2}\R_{\Theta, m})^T}{\sqrt{d}}=\frac{q_{l_1}\R_{\Theta, n-m}k_{l_2}^T}{{\sqrt{d}}}$. BiPE-ALiBi uses ALiBi~\cite{alibi} as the inter-segment encoding. Similarly, the relative segment distance $n-m$ is calculated for token pair $(w_{l_1},w_{l_2})$. The attention score between the two tokens is computed by $\frac{q_{l_1}k_{l_2}^T}{\sqrt{d}} + r|n-m|$, where $r$ is a pre-defined hyper-parameter.

\textbf{Discussion}. The original BERT~\cite{bert} also includes two encodings for representing positions, but its approach differs significantly from BiPE. Primarily, BERT only needs to specify two segments using absolute encoding, tailored for the next sentence prediction task not for length extrapolation. Furthermore, BERT treats a sequence as a flat array of tokens and defines the segments in an arbitrary way, ignoring intrinsic segmentation of language data. See Appendix \ref{appx:more-related-works} for more discussions. An empirical comparison can be found in Section~\ref{sec:exp-ablation}.
\vspace{-5pt}
\subsection{Theoretical Analysis of BiPE}\label{sec:theory}
\vspace{-3pt}
Many previous works are built upon an assumption that tokens are generated in a hierarchical manner in natural language and develop the hierarchical hidden Markov model~\cite{fine1998hierarchical}, hierarchical recurrent model~\cite{chung2016hierarchical}, and hierarchical topic model~\cite{blei2003latent,griffiths2003hierarchical}. We follow to use this assumption to investigate the parameter efficiency of BiPE. In particular, we leverage the (non-deterministic) finite automata (NFA), which is widely used in the field of theoretical computer science. \citet{alur1999communicating} proposed hierarchical finite automata as a practical way to represent such linguistic structures. Inspired by the framework, we introduce a simplified model, Bi-NFA, which restricts the hierarchy level of hierarchical finite automata to two. We compare the parameter efficiency of Transformers to represent NFA and Bi-NFA and show that BiPE has a theoretical advantage over existing positional encoding schemes.

\textbf{NFA.} \looseness=-1A nondeterministic finite automaton (NFA) is a fundamental and essential computational model in computer science \cite{eilenberg1974automata}. An NFA $\gN$ can be defined as a tuple $\gN=(Q,\Sigma,\delta,q_0,F)$, where $Q$ is a set of states, $\Sigma$ is the alphabet of input symbols, $\delta: Q\times\Sigma\rightarrow \gP(Q)$ is a transition function, $q_0\in Q$ is the initial state, and $F\subseteq Q$ is a set of final states. $\gP(Q)$ denotes the power set of $Q$. A string $\mS=[w_1,w_2,\cdots, w_n]\in\Sigma^*$ is accepted by $\gN$ if there exists a sequence of states $r_{0},r_{1},\cdots,r_{n}\in Q$ such that $r_0=q_0$, $r_{i+1}\in\delta(r_i,w_{i+1})$ for $i=0,1,\ldots,n-1$, and $r_n\in F$.

\textbf{Bi-NFA.} We utilize the hierarchical automata to capture the structure of modular segments and introduce the Bi-NFA by restricting the hierarchy level to two. A Bi-NFA is a tuple $\gN=(\mathcal{Q},\Sigma,\delta,q_0, F)$, where $\mathcal{Q}$ is the collection of state sets $Q_1,Q_2,\cdots,Q_n$, $\Sigma$ is the symbol set that includes a segment separator $w^*$, $\delta$ is the transition kernel, $q_0$ is the initial state, and $F$ is the set of accept states. The main difference between Bi-NFA and NFA is that the transitions are constrained by the state sets and the segment separator. Specifically, for any state $q\in Q_i$ and any symbol $w$, we have $\delta(q,w)\subset Q_i$ if $w\neq w^*$ and $\delta(q,w^*)\subset \{q^*_1,\cdots,q^*_n\}$, where $q^*_i$ is the start state in $Q_i$ and $q_0\in\{q^*_1,\cdots,q^*_k\}$. Thus, Bi-NFA stays within the same state set until it reads the segment separator, and then it can move to any other state set in $\mathcal{Q}$. The Bi-NFA can be viewed as a variant of NFA that processes the input sequence segment by segment.

\begin{theorem}[Lower bound for Transformer with absolute positional encoding to represent NFA]
\label{Thm:classic_PE}
    For any size of state set, there exists an NFA $\gN=(Q,\Sigma,\delta,q_0,F)$ such that a Transformer with APE needs at least $O(|Q|^2)$ embedding size to represent the NFA.
\end{theorem}

\begin{theorem}[Upper bound for Transformer with BiPE to represent Bi-NFA]
\label{Thm:BiPE}
   For any Bi-NFA $\gN=(\gQ,\Sigma,\delta,q_0,F)$, $\gQ=\{Q_1,Q_2,\cdots,Q_k\}$ there exists a Transformer with BiPE and $O(k^2+\sum_{i\in[k]} |Q_i|^2)$ embedding size can represent the Bi-NFA.
\end{theorem}
\vspace{-6pt}

The proof of \cref{Thm:classic_PE,Thm:BiPE} can be found in \cref{sec:Proof} and  \cref{Thm:classic_PE} can be extended to relative positional encodings. For a Bi-NFA $\gN=(\gQ,\Sigma,\delta,q_0,F)$, denote $T=\sum_{i\in[k]}|Q_i|$ as the number of states, and assume $|Q_i|=O(\sqrt{T})$. If naively treating $\gN$ as an NFA with $T$ states, we can directly obtain from  \cref{Thm:classic_PE} that an absolute positional encoding-based Transformer requires at least $O(T^2)$ dimensions to represent it. However, from  \cref{Thm:BiPE}, by well exploiting the hierarchical structure, a Transformer with BiPE only requires $O(N^{\frac{3}{2}})$ dimensions. This suggests the superiority of the BiPE over previous method from a theoretical perspective.
   
\vspace{-8pt}
\section{Experiments}\label{sec:exp}
\vspace{-2pt}
In this section, we empirically study the effectiveness of our BiPE method. In particular, we aim to answer the following questions through experiments:
\vspace{-6pt}
\begin{itemize}
    \item \textbf{Q1}: Do the theoretical results regarding parameter efficiency of BiPE hold in practice? 
    (Sec \ref{sec:exp-parameter-efficiency})
    \item \textbf{Q2}: Does BiPE bring superior length extrapolation capabilities in real-world tasks? 
    (Sec \ref{sec:lm})
    \item \textbf{Q3}: Does BiPE help Transformer-based language models better understand long text? 
    (Sec \ref{sec:exp-long-text})
    \item \textbf{Q4}: Does BiPE hurt performance on normal-length text? 
    (Sec \ref{sec:exp-short-text})
    \item \textbf{Q5}: Is each design choice in BiPE helpful? 
    (Sec \ref{sec:exp-ablation})
\end{itemize}
\vspace{-6pt}
We will thoroughly answer each question with carefully designed experiments on widely used benchmarks as below. We also cover different modalities in the experiments, including math (arithmetical reasoning task in Section \ref{sec:exp-parameter-efficiency}), natural language (PG19\&ArXiv task in Section \ref{sec:lm}) and code (Github task Section \ref{sec:lm}). We run each experiment multiple times with different random seeds and report the averaged results. Due to space limits, we present more details and additional results in Appendix \ref{app_exp:add_exp}.

\vspace{-3pt}
\subsection{Capacity Experiments}\label{sec:exp-parameter-efficiency}

\begin{figure}[t]
\centering
\includegraphics[width=0.95\linewidth]{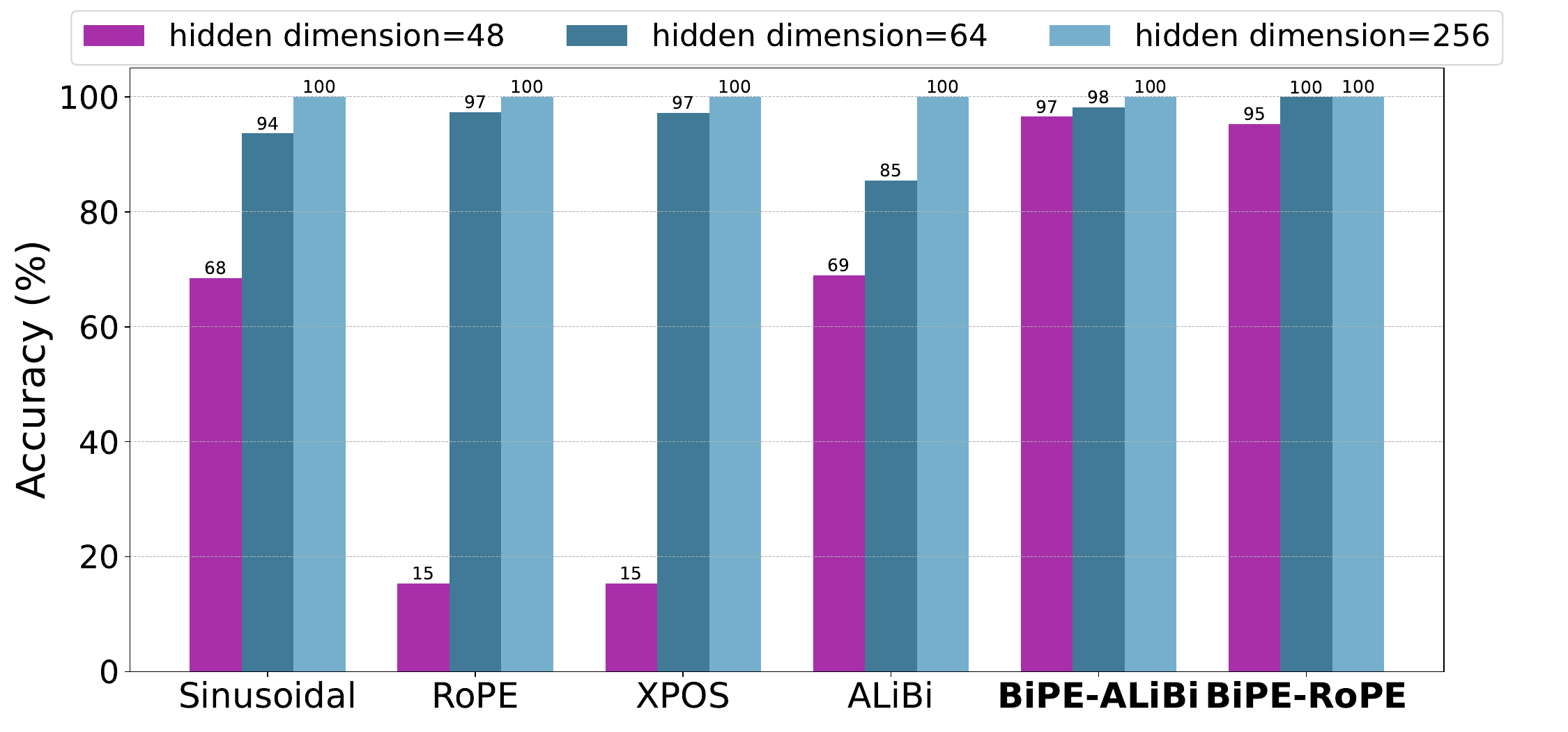}
\vspace{-4pt}
\caption{
Accuracy of Transformer models with different positional encoding methods on the Arithmetic task. Our BiPE method consistently performs best on different scales of parameters.
}
\label{fig:4_math}
\vspace{-14pt}
\end{figure}
\textbf{Tasks}. To empirically verify the parameter efficiency brought by our BiPE method, we conduct experiments on the Arithmetic task \citep{feng2023towards}, which is recently used as a proxy to examine the mathematical reasoning capability of language models. Given an arithmetical expression consisting of numbers, basic operations ($+,-,\times,\div,=$) and brackets, e.g., $(1+2)\times(3+5)=$, this task requires language models to calculate and generate the correct result, e.g., $24$. Following \citet{feng2023towards}, we train all models using Chain-of-Thought demonstrations (See Appendix \ref{app:cot_math}). The evaluation metric is the accuracy of the final answer.

\textbf{Settings}. The Arithmetic dataset from~\citet{feng2023towards} consists of 1 million training samples and 100k test samples in total. We choose the standard decoder-only Transformer language model as the base model and compare our BiPE method with the following competitive positional encodings: 1) Sinusoidal PE~\cite{vaswani2017attention}; 2) RoPE~\cite{rope}; 3) XPOS~\cite{sun2022length}; 4) ALiBi~\cite{alibi}. In particular, we implement two versions of our BiPE, BiPE-RoPE and BiPE-ALiBi, which instantiates the inter-segment encoding via RoPE and ALiBi respectively. The segment boundary is simply determined by the equal sign ``$=$''. Following~\citet{feng2023towards}, we set the number of layers to 3 and the number of attention heads to 4. To evaluate the parameter efficiency, we vary the hidden dimension in $[48, 64, 256]$. Additional experimental details are presented in Appendix~\ref{app:cot_math}.

\textbf{Results}. In Figure \ref{fig:4_math}, it can be easily seen that given a similar amount of parameters, BiPE-based language models consistently outperform other baselines on this task. For example, when the hidden dimension is 48, other positional encoding methods achieve inferior accuracy (below 70\%), while BiPE-ALiBi and BiPE-RoPE achieve high accuracy of 97\% and 95\% respectively. This result indeed well aligns with our theoretical results in Section \ref{sec:theory}, which further serves as a strong support for the bilevel design of our BiPE.

\begin{figure*}[t]
    \begin{center}
    \includegraphics[width=1.0\linewidth]{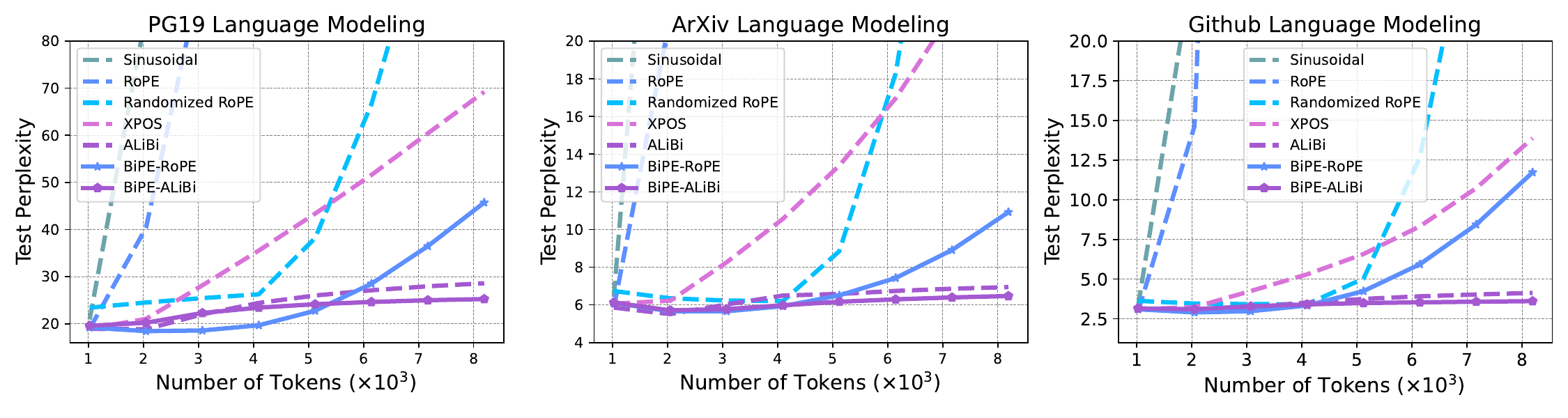}
    \end{center}
    \vspace{-1.0em}
    \caption{Language modeling perplexity with varying evaluation sequence lengths for models trained on sequence length 1024.}
    \label{fig:lm_ppl}
    \vspace{-12pt}
\end{figure*}

\vspace{-3pt}
\subsection{Length Extrapolation Experiments}\label{sec:lm}
\textbf{Tasks}. We test the length extrapolation capability of Transformer-based language models with different positional encoding methods. Following~\citet{chi2022kerple}, we use the Pile~\cite{gao2020pile} dataset as the pre-training corpus and evaluate the log perplexity of pre-trained language models on the test set of PG19~\cite{pg19}, arXiv and Github~\cite{gao2020pile}. We conduct the non-overlapping evaluation when computing the perplexity score.

\looseness=-1\textbf{Settings}. We set the pre-training sequence length to 1024, and evaluate the zero-shot perplexity on sequence lengths $[1024, 2048, 3072, 4096, 5120, 6144, 7168, 8192]$ on downstream datasets. We choose the standard decoder-only Transformer as the base model and compare our BiPE methods (BiPE-RoPE and BiPE-ALiBi) with the following positional encodings: 1) Sinusoidal PE~\cite{vaswani2017attention}; 2) RoPE~\cite{rope}; 3) Randomized RoPE~\cite{randompos}; 4) XPOS~\cite{sun2022length}; 5) ALiBi~\cite{alibi}. The segment boundary is determined by full stop ``\texttt{.}'' and newline ``\texttt{\textbackslash n}'' for general purposes. For the Transformer-based language model, we set the number of layers to 12, the hidden dimension to 768, and the number of attention heads to 12. The total number of model parameters is approximately 155M. Additional experimental details are presented in Appendix~\ref{app:extra_exp}.

\looseness=-1\textbf{Results}. The results are presented in Figure~\ref{fig:lm_ppl}. Our BiPE methods achieve consistently superior performance on sequences with lengths larger than the training length. For example, our BiPE-ALiBi outperforms its counterpart ALiBi, which is also the best baseline method, by 3.35 points (25.24 v.s. 28.59 perplexity) on PG19 with 8192 sequence length. Compared to RoPE which performs well on sequences with the in-distribution length but yields a significant performance drop on longer sequences, our BiPE method substantially improves its length extrapolation capabilities, e.g., 19.67 v.s. 158 perplexity on PG19 with the 4096 sequence length. Notably, the benefit brought by our BiPE is also consistent across all three evaluation datasets covering text data in different modalities, underscoring the better length extrapolation capability of our BiPE in real-world tasks.
\begin{figure*}[t]
\centering
\includegraphics[width=1.0\linewidth]{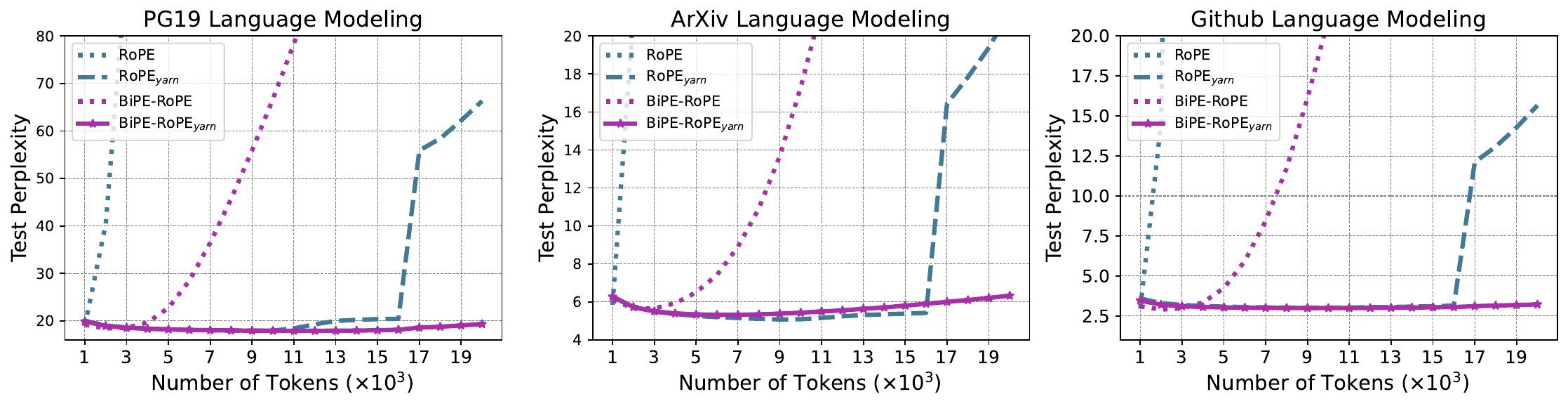}
\vspace{-1em}
\caption{
Language modeling perplexity with varying evaluation sequence lengths for RoPE and BiPE-RoPE finetuned with YaRN.
}
\label{fig:4_yarn}
\vspace{-8pt}
\end{figure*}

\textbf{Integrating BiPE with fine-tuning strategies}. One line of recent improvements on length extrapolation comes from continued fine-tuning RoPE-based language models with Position Interpolation techniques~\cite{chen2023extending, peng2023yarn}. To further investigate the compatibility of our BiPE method with Position Interpolation, we use YaRN ~\cite{peng2023yarn} to finetune the language model pre-trained on the Pile dataset with RoPE~\citep{rope} and our BiPE-RoPE, and check the improvements on downstream datasets. The results are presented in Figure~\ref{fig:4_yarn}. Similar to the zero-shot evaluation setting, our BiPE-RoPE achieves consistently better performance on longer sequences compared to RoPE after finetuning. Furthermore, although YaRN improves the length extrapolation capability of RoPE to some extent, it still suffers from performance drop when being evaluated on very long sequences, e.g., 11k/16k/16k for PG19/ArXiv/Github. In contrast, our BiPE-RoPE combined with YaRN yields much better length extrapolation capability, i.e. maintaining a consistently low perplexity across sequences with lengths up to 20k. Please refer to Appendix~\ref{app_exp:yarn} for more experimental details.
\vspace{-6pt}
\subsection{Long Context Benchmark}\label{sec:exp-long-text}
\begin{table*}[t]
\caption{Performance comparison on SCROLLS benchmark. Abbreviations for dataset names: Qasper (Qas), ContractNLI (CNLI), QMSum (QMS), NarrativeQA (NQA), SummScreenFD (SumS), GovReport (GovR), and QuALITY (QuAL). Rgm denotes the geometric mean of ROUGE-1,2, L. The statistics of median sequence lengths are from \citet{ainslie2023colt, li2023functional}. Best performing results are highlighted in \textbf{bold}.}
\label{tab:scrolls}
\centering
\begin{tabular}{@{}lcccccccc@{}}
\toprule
 & \textbf{QAS} & \textbf{CNLI} & \textbf{QMS} & \textbf{NQA} & \textbf{SumS} & \textbf{GovR} & \textbf{QuAL} & \textbf{Average} \\ \midrule
\textbf{Metric} & F1 & EM & Rgm & F1 & Rgm & Rgm & EM &  \\
\textbf{Median length} &	5472&	2148&	14197&	57829&	9046&	8841&	7171\\\midrule
Sinusoidal & 9.3 & 57.7 & \textbf{12.42} & 10.1  & 7.46 & 12.49 & 1.9 & 15.89 \\
Randomized RoPE & 12.3 & 52.4 & 11.80 & 10.8 & 7.19 & 18.95 & \textbf{10.4} & 17.71 \\
\midrule
ALiBi & 12.0 & 68.8 & 10.27 & 3.2 & 6.00 & 23.14 & 0.0 & 17.62 \\
BiPE-ALiBi & 12.7 & 67.8 & 10.44 & 2.6 & 7.89  & 27.52 & 0.0 & 18.34 \\
\midrule
RoPE & 16.4 & 67.8 & 10.13 & 9.7  & \textbf{9.88} & 14.33 & 0.4 & 18.38  \\
BiPE-RoPE & \textbf{21.2} & \textbf{68.9} & 10.64 & \textbf{12.3} & 8.13 & \textbf{27.92} & 7.4 & \textbf{22.36}\\
\midrule
\midrule
RoPE$_{yarn}$ & 14.7 & 66.9 & 9.04 & 12.2 & 8.48 & 27.56 & \textbf{22.2} & 23.01 \\
BiPE-RoPE$_{yarn}$ & \textbf{20.9} & \textbf{69.0} & \textbf{10.57} & \textbf{13.3} & \textbf{9.40} & \textbf{28.31} & 20.3 & \textbf{24.53}\\

\bottomrule     
\end{tabular}
\vspace{-6pt}
\end{table*}

\textbf{Tasks and settings}. 
\looseness=-1To evaluate the model's performance of long context understanding, we further fine-tune the pre-trained checkpoints on SCROLLS~\cite{shaham2022scrolls}, a long text benchmark that consists of seven distinct datasets covering different tasks. Following ~\citet{shaham2022scrolls,ainslie2023colt}, we use three evaluation metrics for different tasks: Rgm score (the geometric mean of ROUGE-1,2,L), unigram overlap (F1) and exact match (EM). The average score across different datasets is also reported. We finetune pre-trained models using a sequence length of 8192 and select the model checkpoint that achieves the best performance on the validation set for the final evaluation. The test results are obtained from the official SCROLLS website. Additional experimental details are presented in Appendix~\ref{app_exp:long_context_benchmark}.

\textbf{Results}. The empirical results are provided in Table~\ref{tab:scrolls}. First, BiPE-RoPE and BiPE-ALiBi exhibit better performance than RoPE and ALiBi, respectively. For example, our BiPE-RoPE outperforms its counterpart RoPE, which is also the best baseline method, by 3.98 points (22.36 v.s. 18.38 average score). Besides, BiPE-RoPE achieves the highest average score, surpassing other methods by a margin of over 3 points. On a task-by-task basis, BiPE-RoPE achieves the top score in 4 out of the 7 tasks. We also compare the two YaRN-finetuned models, i.e., BiPE-RoPE$_{yarn}$ and RoPE$_{yarn}$. We can see that BiPE-RoPE$_{yarn}$ still consistently outperforms RoPE$_{yarn}$ across 6 of 7 tasks and achieves a better average score. The results strengthen the effectiveness of BiPE in long-context modeling.

\textbf{Discussions.} We can also observe that the performance gap between RoPE and BiPE-RoPE is more significant than that between ALiBi and BiPE-ALiBi. We hypothesize that this phenomenon is due to the design differences between ALiBi and RoPE. ALiBi incorporates relative positional information as an additive term in the attention module with an exponential decay rate as the relative distance increases. Using ALiBi on segment indexes (BiPE-ALiBi) will still bias the attention module towards local attention~\cite{chi2023dissecting}. Thus, the performance gap is not substantial. Different from ALiBi, RoPE rotates the query and key vectors and allows the context to determine the positional correlations. Therefore, the change between BiPE-RoPE and RoPE is more significant, leading to a larger performance gap.

\subsection{Normal-length Benchmark}\label{sec:exp-short-text}
\begin{table*}[t]
\centering
\caption{Zero-shot and few-shot performance on standard benchmarks. BiPE-based models perform on par with other methods.}
\begin{tabular}{lcccccc}
\toprule
\multirow{2.5}{*}{\textbf{Model}} & \multicolumn{4}{c}{\textbf{Zero-Shot}} & \multicolumn{2}{c}{\textbf{Few-Shot}}\\ \cmidrule(r){2-5} \cmidrule(r){6-7} & RACE & WinoGrande & TruthfulQA & PIQA & HellaSwag & MMLU \\ \midrule
Sinusoidal & 27.85 & 52.09 & 45.92 & 60.88 & 29.70 & 26.49 \\
Randomized RoPE & 26.41 & 50.99 & 45.53 & 60.66 & 29.30 & 25.20 \\
XPOS & 27.56 & 52.96 & 45.22 & 60.88 & 30.86 & 25.96 \\
\midrule
ALiBi & 27.08 & 52.72 & 46.24 & 60.50 & 31.47 & 26.49 \\ 
BiPE-ALiBi & 28.42 & 49.25 & 45.79 & 60.72 & 30.60  & 25.74 \\ 
\midrule
RoPE & 29.00 & 51.54 & 44.67 & 60.66 & 30.86 & 26.43 \\ 
BiPE-RoPE & 28.04 & 52.01 & 45.64 & 59.74 & 30.93 & 26.91 \\ 
\midrule
\midrule
RoPE$_{yarn}$  & 27.08 & 53.35  & 45.69  & 60.12  & 30.52  & 26.16  \\ 
BiPE-RoPE$_{yarn}$ & 27.56  & 51.38 & 45.80  & 60.72  & 30.61  & 26.22  \\ 
\bottomrule
\end{tabular}
\label{tab:zero_few}
\end{table*}

\textbf{Tasks and settings}. 
In this experiment, we evaluate the zero-shot and few-shot performance~\cite{eval-harness} of pre-trained models on a range of ``in-distribution'' benchmark tasks where the sequence length is normal. In particular, we use RACE~\cite{lai-etal-2017-race}, WinoGrade~\cite{sakaguchi2020winogrande}, TruthfulQA mc2~\cite{lin-etal-2022-truthfulqa}, and PIQA~\cite{gao2020pile} benchmarks for the zero-shot evaluation, and employ 10-shot HellaSwag~\cite{zellers2019hellaswag} and 5-shot MMLU~\cite{mmlu} for the few-shot evaluation. The evaluation metrics are task-specific: for RACE, WinoGrande, TruthfulQA, PIQA, and MMLU, we report accuracy; and for HellaSwag, we report normalized accuracy.

\textbf{Results}. The empirical results are provided
in Table~\ref{tab:zero_few}. It can be easily seen that BiPE-RoPE and BiPE-ALiBi achieve comparable performance with other positional encoding methods on sequences with in-distribution lengths, which demonstrates that our BiPE methods achieve better length extrapolation performance without sacrificing in-distribution performance.

\vspace{-2pt}
\subsection{Ablation Study}\label{sec:exp-ablation}
\vspace{-2pt}

\begin{figure*}[t]
\centering
\includegraphics[width=0.8\linewidth]{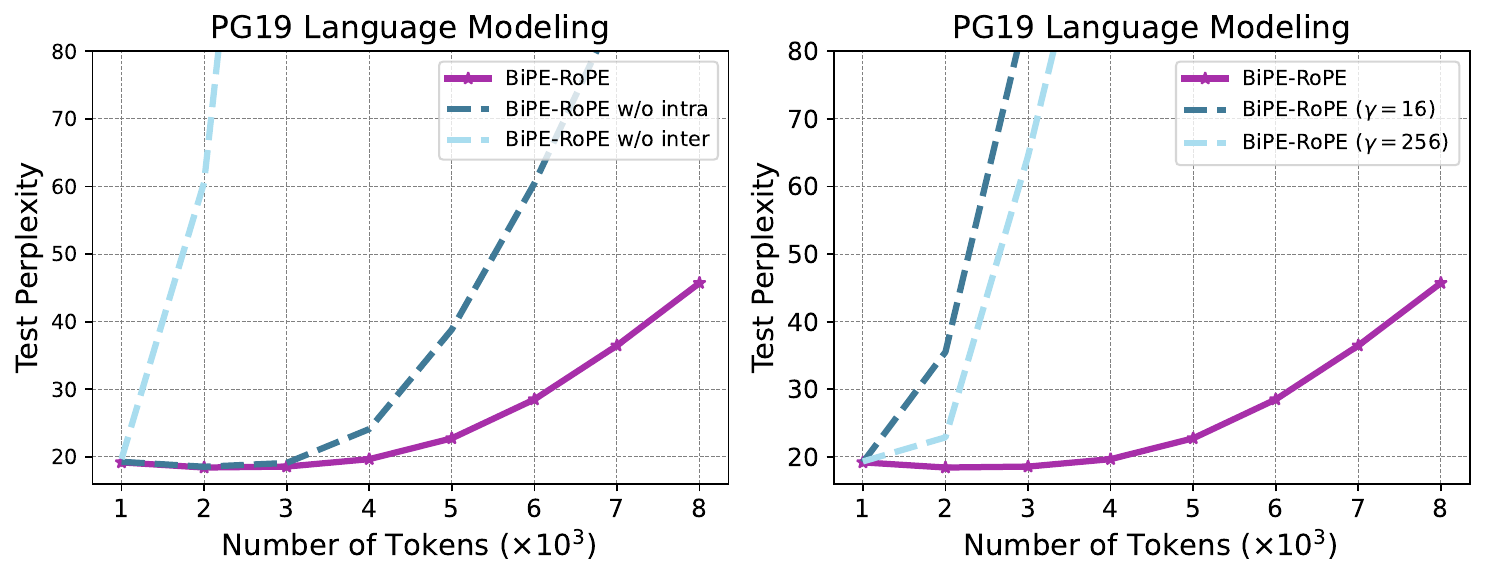}
\vspace{-0.5em}
\caption{
\textbf{Left:} Language modeling perplexity with varying evaluation sequence lengths for BiPE-RoPE without intra-segment encoding or inter-segment encoding on PG19 dataset. \textbf{Right:} Language modeling perplexity with varying evaluation sequence lengths for BiPE-RoPE with fixed segment lengths on PG19 dataset.
}
\vspace{-10pt}
\label{fig:5_abalation}
\end{figure*}

\textbf{Effectiveness of each positional encoding}. BiPE leverages two positional encodings, with one corresponding to the token index within each segment (intra-segment encoding) and another for the segment index (inter-segment encoding). To check their effectiveness, we conducted an experiment where we removed one encoding at a time to assess the impact on model performance. We follow the same pre-training setting in Section~\ref{sec:lm} and evaluate perplexity on PG19. In Figure~\ref{fig:5_abalation} left, the degradation in performance is observed upon the removal of either encoding, which clearly demonstrates that both of them are important.

\looseness=-1\textbf{Segmentation choices}. In Section~\ref{sec:lm}, we use full stop ``\texttt{.}'' and newline ``\texttt{\textbackslash n}'' for text segmentation. One may wonder whether using a fixed segment length instead of pre-defined symbols works in practice. To check this, we assume each length is constant $\gamma$, pre-train BiPE language models and evaluate the performance on the PG19 dataset. In Figure~\ref{fig:5_abalation} right, we can see that this naive approach does not achieve the same level of performance as using natural segmentation.

\vspace{-4pt}
\section{Conclusion and Future Directions}
In this paper, we introduce BiPE, a novel bilevel positional encoding scheme designed to improve length extrapolation. For each position, our BiPE combines 1) an intra-segment encoding that identifies the location within its segment via APE, and 2) an inter-segment encoding that specifies the segment to which it belongs via RPE. The intra-segment encoding assists the model in capturing the semantic information within each segment and and the inter-segment encoding models the relationships between segments. This bilevel design well aligns with the intrinsic segmentation of text data and enhances length extrapolation. Our BiPE is further supported by theoretical analysis of its expressiveness. All experiments verify the length extrapolation capability of our BiPE across tasks of different text modalities.

There are also several future directions worth investigating. First, the intrinsic segmentation of text data yields a hierarchical structure, e.g., sentences$\to$paragraphs$\to$documents. It would be beneficial to confirm whether expanding our bilevel design to a hierarchical version results in improved length extrapolation. Second, there exist sequence data that do not have clear boundary of segmentations, e.g., time series, amino acid and gene sequence. Future research could explore better and more comprehensive segmentation methods for general purposes with our BiPE method.
\newpage

\section*{Impact Statement}

This paper presents work whose goal is to advance the field of Machine Learning. There are many potential societal consequences of our work, none which we feel must be specifically highlighted here.

\section*{Acknowledgement}
Di He is supported by National Key R\&D Program of China (2022ZD0160300) and National Science Foundation of China (NSFC62376007). Liwei Wang is supported by National Science Foundation of China (NSFC62276005).


\bibliography{main}

\begin{thebibliography}{61}
\providecommand{\natexlab}[1]{#1}
\providecommand{\url}[1]{\texttt{#1}}
\expandafter\ifx\csname urlstyle\endcsname\relax
  \providecommand{\doi}[1]{doi: #1}\else
  \providecommand{\doi}{doi: \begingroup \urlstyle{rm}\Url}\fi

\bibitem[Ainslie et~al.(2023)Ainslie, Lei, de~Jong, Ontanon, Brahma, Zemlyanskiy, Uthus, Guo, Lee-Thorp, Tay, Sung, and Sanghai]{ainslie2023colt}
Ainslie, J., Lei, T., de~Jong, M., Ontanon, S., Brahma, S., Zemlyanskiy, Y., Uthus, D., Guo, M., Lee-Thorp, J., Tay, Y., Sung, Y.-H., and Sanghai, S.
\newblock Co{LT}5: Faster long-range transformers with conditional computation.
\newblock In \emph{The 2023 Conference on Empirical Methods in Natural Language Processing}, 2023.

\bibitem[Alur et~al.(1999)Alur, Kannan, and Yannakakis]{alur1999communicating}
Alur, R., Kannan, S., and Yannakakis, M.
\newblock Communicating hierarchical state machines.
\newblock In \emph{Automata, Languages and Programming: 26th International Colloquium, ICALP’99 Prague, Czech Republic, July 11--15, 1999 Proceedings 26}, pp.\  169--178. Springer, 1999.

\bibitem[Anil et~al.(2022)Anil, Wu, Andreassen, Lewkowycz, Misra, Ramasesh, Slone, Gur-Ari, Dyer, and Neyshabur]{anil2022exploring}
Anil, C., Wu, Y., Andreassen, A., Lewkowycz, A., Misra, V., Ramasesh, V., Slone, A., Gur-Ari, G., Dyer, E., and Neyshabur, B.
\newblock Exploring length generalization in large language models.
\newblock \emph{Advances in Neural Information Processing Systems}, 35:\penalty0 38546--38556, 2022.

\bibitem[Bai et~al.(2021)Bai, Shi, Lin, Xie, Tan, Xiong, Gao, and Li]{bai2021segatron}
Bai, H., Shi, P., Lin, J., Xie, Y., Tan, L., Xiong, K., Gao, W., and Li, M.
\newblock Segatron: Segment-aware transformer for language modeling and understanding.
\newblock In \emph{Proceedings of the AAAI Conference on Artificial Intelligence}, volume~35, pp.\  12526--12534, 2021.

\bibitem[Blei et~al.(2003)Blei, Ng, and Jordan]{blei2003latent}
Blei, D.~M., Ng, A.~Y., and Jordan, M.~I.
\newblock Latent dirichlet allocation.
\newblock \emph{Journal of machine Learning research}, 3\penalty0 (Jan):\penalty0 993--1022, 2003.

\bibitem[Brown et~al.(2020)Brown, Mann, Ryder, Subbiah, Kaplan, Dhariwal, Neelakantan, Shyam, Sastry, Askell, et~al.]{brown2020language}
Brown, T., Mann, B., Ryder, N., Subbiah, M., Kaplan, J.~D., Dhariwal, P., Neelakantan, A., Shyam, P., Sastry, G., Askell, A., et~al.
\newblock Language models are few-shot learners.
\newblock \emph{Advances in neural information processing systems}, 33:\penalty0 1877--1901, 2020.

\bibitem[Chen et~al.(2023{\natexlab{a}})Chen, Li, Meng, Liang, and Bing]{chen2023clex}
Chen, G., Li, X., Meng, Z., Liang, S., and Bing, L.
\newblock Clex: Continuous length extrapolation for large language models.
\newblock \emph{arXiv preprint arXiv:2310.16450}, 2023{\natexlab{a}}.

\bibitem[Chen et~al.(2022)Chen, Chu, Wiseman, and Gimpel]{chen-etal-2022-summscreen}
Chen, M., Chu, Z., Wiseman, S., and Gimpel, K.
\newblock {S}umm{S}creen: A dataset for abstractive screenplay summarization.
\newblock In \emph{Proceedings of the 60th Annual Meeting of the Association for Computational Linguistics (Volume 1: Long Papers)}, pp.\  8602--8615, May 2022.

\bibitem[Chen et~al.(2023{\natexlab{b}})Chen, Wong, Chen, and Tian]{chen2023extending}
Chen, S., Wong, S., Chen, L., and Tian, Y.
\newblock Extending context window of large language models via positional interpolation.
\newblock \emph{arXiv preprint arXiv:2306.15595}, 2023{\natexlab{b}}.

\bibitem[Chi et~al.(2022)Chi, Fan, Ramadge, and Rudnicky]{chi2022kerple}
Chi, T.-C., Fan, T.-H., Ramadge, P.~J., and Rudnicky, A.
\newblock Kerple: Kernelized relative positional embedding for length extrapolation.
\newblock \emph{Advances in Neural Information Processing Systems}, 35:\penalty0 8386--8399, 2022.

\bibitem[Chi et~al.(2023)Chi, Fan, Rudnicky, and Ramadge]{chi2023dissecting}
Chi, T.-C., Fan, T.-H., Rudnicky, A., and Ramadge, P.
\newblock Dissecting transformer length extrapolation via the lens of receptive field analysis.
\newblock In \emph{Proceedings of the 61st Annual Meeting of the Association for Computational Linguistics (Volume 1: Long Papers)}, pp.\  13522--13537, 2023.

\bibitem[Chowdhery et~al.(2022)Chowdhery, Narang, Devlin, Bosma, Mishra, Roberts, Barham, Chung, Sutton, Gehrmann, Schuh, Shi, Tsvyashchenko, Maynez, Rao, Barnes, Tay, Shazeer, Prabhakaran, Reif, Du, Hutchinson, Pope, Bradbury, Austin, Isard, Gur-Ari, Yin, Duke, Levskaya, Ghemawat, Dev, Michalewski, Garcia, Misra, Robinson, Fedus, Zhou, Ippolito, Luan, Lim, Zoph, Spiridonov, Sepassi, Dohan, Agrawal, Omernick, Dai, Pillai, Pellat, Lewkowycz, Moreira, Child, Polozov, Lee, Zhou, Wang, Saeta, Diaz, Firat, Catasta, Wei, Meier-Hellstern, Eck, Dean, Petrov, and Fiedel]{chowdhery2022palm}
Chowdhery, A., Narang, S., Devlin, J., Bosma, M., Mishra, G., Roberts, A., Barham, P., Chung, H.~W., Sutton, C., Gehrmann, S., Schuh, P., Shi, K., Tsvyashchenko, S., Maynez, J., Rao, A., Barnes, P., Tay, Y., Shazeer, N., Prabhakaran, V., Reif, E., Du, N., Hutchinson, B., Pope, R., Bradbury, J., Austin, J., Isard, M., Gur-Ari, G., Yin, P., Duke, T., Levskaya, A., Ghemawat, S., Dev, S., Michalewski, H., Garcia, X., Misra, V., Robinson, K., Fedus, L., Zhou, D., Ippolito, D., Luan, D., Lim, H., Zoph, B., Spiridonov, A., Sepassi, R., Dohan, D., Agrawal, S., Omernick, M., Dai, A.~M., Pillai, T.~S., Pellat, M., Lewkowycz, A., Moreira, E., Child, R., Polozov, O., Lee, K., Zhou, Z., Wang, X., Saeta, B., Diaz, M., Firat, O., Catasta, M., Wei, J., Meier-Hellstern, K., Eck, D., Dean, J., Petrov, S., and Fiedel, N.
\newblock Palm: Scaling language modeling with pathways, 2022.

\bibitem[Chowdhury \& Caragea(2023)Chowdhury and Caragea]{chowdhury2023monotonic}
Chowdhury, J.~R. and Caragea, C.
\newblock Monotonic location attention for length generalization.
\newblock \emph{arXiv preprint arXiv:2305.20019}, 2023.

\bibitem[Chung et~al.(2016)Chung, Ahn, and Bengio]{chung2016hierarchical}
Chung, J., Ahn, S., and Bengio, Y.
\newblock Hierarchical multiscale recurrent neural networks.
\newblock \emph{arXiv preprint arXiv:1609.01704}, 2016.

\bibitem[Dasigi et~al.(2021)Dasigi, Lo, Beltagy, Cohan, Smith, and Gardner]{dasigi2021qasper}
Dasigi, P., Lo, K., Beltagy, I., Cohan, A., Smith, N.~A., and Gardner, M.
\newblock A dataset of information-seeking questions and answers anchored in research papers.
\newblock In \emph{Proceedings of the 2021 Conference of the North American Chapter of the Association for Computational Linguistics: Human Language Technologies}, June 2021.

\bibitem[Devlin et~al.(2019{\natexlab{a}})Devlin, Chang, Lee, and Toutanova]{bert}
Devlin, J., Chang, M.-W., Lee, K., and Toutanova, K.
\newblock {BERT}: Pre-training of deep bidirectional transformers for language understanding.
\newblock In \emph{Proceedings of the 2019 Conference of the North {A}merican Chapter of the Association for Computational Linguistics: Human Language Technologies, Volume 1 (Long and Short Papers)}, June 2019{\natexlab{a}}.

\bibitem[Devlin et~al.(2019{\natexlab{b}})Devlin, Chang, Lee, and Toutanova]{devlin-etal-2019-bert}
Devlin, J., Chang, M.-W., Lee, K., and Toutanova, K.
\newblock {BERT}: Pre-training of deep bidirectional transformers for language understanding.
\newblock In \emph{Proceedings of the 2019 Conference of the North {A}merican Chapter of the Association for Computational Linguistics: Human Language Technologies, Volume 1 (Long and Short Papers)}, pp.\  4171--4186, June 2019{\natexlab{b}}.

\bibitem[Eilenberg(1974)]{eilenberg1974automata}
Eilenberg, S.
\newblock \emph{Automata, languages, and machines}.
\newblock Academic press, 1974.

\bibitem[Feng et~al.(2023)Feng, Zhang, Gu, Ye, He, and Wang]{feng2023towards}
Feng, G., Zhang, B., Gu, Y., Ye, H., He, D., and Wang, L.
\newblock Towards revealing the mystery behind chain of thought: A theoretical perspective.
\newblock In \emph{Thirty-seventh Conference on Neural Information Processing Systems}, 2023.

\bibitem[Fine et~al.(1998)Fine, Singer, and Tishby]{fine1998hierarchical}
Fine, S., Singer, Y., and Tishby, N.
\newblock The hierarchical hidden markov model: Analysis and applications.
\newblock \emph{Machine learning}, 32:\penalty0 41--62, 1998.

\bibitem[Gao et~al.(2020)Gao, Biderman, Black, Golding, Hoppe, Foster, Phang, He, Thite, Nabeshima, et~al.]{gao2020pile}
Gao, L., Biderman, S., Black, S., Golding, L., Hoppe, T., Foster, C., Phang, J., He, H., Thite, A., Nabeshima, N., et~al.
\newblock The {P}ile: An 800{GB} dataset of diverse text for language modeling.
\newblock \emph{arXiv preprint arXiv:2101.00027}, 2020.

\bibitem[Gao et~al.(2023)Gao, Tow, Abbasi, Biderman, Black, DiPofi, Foster, Golding, Hsu, Le~Noac'h, Li, McDonell, Muennighoff, Ociepa, Phang, Reynolds, Schoelkopf, Skowron, Sutawika, Tang, Thite, Wang, Wang, and Zou]{eval-harness}
Gao, L., Tow, J., Abbasi, B., Biderman, S., Black, S., DiPofi, A., Foster, C., Golding, L., Hsu, J., Le~Noac'h, A., Li, H., McDonell, K., Muennighoff, N., Ociepa, C., Phang, J., Reynolds, L., Schoelkopf, H., Skowron, A., Sutawika, L., Tang, E., Thite, A., Wang, B., Wang, K., and Zou, A.
\newblock A framework for few-shot language model evaluation, 12 2023.
\newblock URL \url{https://zenodo.org/records/10256836}.

\bibitem[Griffiths et~al.(2003)Griffiths, Jordan, Tenenbaum, and Blei]{griffiths2003hierarchical}
Griffiths, T., Jordan, M., Tenenbaum, J., and Blei, D.
\newblock Hierarchical topic models and the nested chinese restaurant process.
\newblock \emph{Advances in neural information processing systems}, 16, 2003.

\bibitem[Halliday \& Matthiessen(2013)Halliday and Matthiessen]{halliday2013halliday}
Halliday, M. A.~K. and Matthiessen, C.~M.
\newblock \emph{Halliday's introduction to functional grammar}.
\newblock Routledge, 2013.

\bibitem[Han et~al.(2023)Han, Wang, Xiong, Chen, Ji, and Wang]{han2023lminfinite}
Han, C., Wang, Q., Xiong, W., Chen, Y., Ji, H., and Wang, S.
\newblock Lm-infinite: Simple on-the-fly length generalization for large language models, 2023.

\bibitem[Haviv et~al.(2022)Haviv, Ram, Press, Izsak, and Levy]{haviv-etal-2022-transformer}
Haviv, A., Ram, O., Press, O., Izsak, P., and Levy, O.
\newblock Transformer language models without positional encodings still learn positional information.
\newblock In \emph{Findings of the Association for Computational Linguistics: EMNLP 2022}, pp.\  1382--1390, December 2022.

\bibitem[Hendrycks et~al.(2021)Hendrycks, Burns, Basart, Zou, Mazeika, Song, and Steinhardt]{mmlu}
Hendrycks, D., Burns, C., Basart, S., Zou, A., Mazeika, M., Song, D., and Steinhardt, J.
\newblock Measuring massive multitask language understanding.
\newblock \emph{Proceedings of the International Conference on Learning Representations (ICLR)}, 2021.

\bibitem[Huang et~al.(2021)Huang, Cao, Parulian, Ji, and Wang]{huang2021govreport}
Huang, L., Cao, S., Parulian, N., Ji, H., and Wang, L.
\newblock Efficient attentions for long document summarization.
\newblock In \emph{Proceedings of the 2021 Conference of the North American Chapter of the Association for Computational Linguistics: Human Language Technologies}, pp.\  1419--1436, June 2021.

\bibitem[Jin et~al.(2024)Jin, Han, Yang, Jiang, Liu, Chang, Chen, and Hu]{jin2024llm}
Jin, H., Han, X., Yang, J., Jiang, Z., Liu, Z., Chang, C.-Y., Chen, H., and Hu, X.
\newblock Llm maybe longlm: Self-extend llm context window without tuning.
\newblock \emph{arXiv preprint arXiv:2401.01325}, 2024.

\bibitem[Kazemnejad et~al.(2023)Kazemnejad, Padhi, Ramamurthy, Das, and Reddy]{kazemnejad2023impact}
Kazemnejad, A., Padhi, I., Ramamurthy, K.~N., Das, P., and Reddy, S.
\newblock The impact of positional encoding on length generalization in transformers.
\newblock \emph{arXiv preprint arXiv:2305.19466}, 2023.

\bibitem[Ko{\v{c}}isk{\'y} et~al.(2018)Ko{\v{c}}isk{\'y}, Schwarz, Blunsom, Dyer, Hermann, Melis, and Grefenstette]{kocisky2018narrativeqa}
Ko{\v{c}}isk{\'y}, T., Schwarz, J., Blunsom, P., Dyer, C., Hermann, K.~M., Melis, G., and Grefenstette, E.
\newblock The {N}arrative{QA} reading comprehension challenge.
\newblock \emph{Transactions of the Association for Computational Linguistics}, 6, 2018.

\bibitem[Koreeda \& Manning(2021)Koreeda and Manning]{koreeda-manning-2021-contractnli-dataset}
Koreeda, Y. and Manning, C.
\newblock {C}ontract{NLI}: A dataset for document-level natural language inference for contracts.
\newblock In \emph{Findings of the Association for Computational Linguistics: EMNLP 2021}, pp.\  1907--1919, November 2021.

\bibitem[Lai et~al.(2017)Lai, Xie, Liu, Yang, and Hovy]{lai-etal-2017-race}
Lai, G., Xie, Q., Liu, H., Yang, Y., and Hovy, E.
\newblock {RACE}: Large-scale {R}e{A}ding comprehension dataset from examinations.
\newblock In \emph{Proceedings of the 2017 Conference on Empirical Methods in Natural Language Processing}, pp.\  785--794, September 2017.

\bibitem[Li et~al.(2023)Li, You, Guruganesh, Ainslie, Ontanon, Zaheer, Sanghai, Yang, Kumar, and Bhojanapalli]{li2023functional}
Li, S., You, C., Guruganesh, G., Ainslie, J., Ontanon, S., Zaheer, M., Sanghai, S., Yang, Y., Kumar, S., and Bhojanapalli, S.
\newblock Functional interpolation for relative positions improves long context transformers.
\newblock \emph{arXiv preprint arXiv:2310.04418}, 2023.

\bibitem[Lightman et~al.(2023)Lightman, Kosaraju, Burda, Edwards, Baker, Lee, Leike, Schulman, Sutskever, and Cobbe]{lightman2023lets}
Lightman, H., Kosaraju, V., Burda, Y., Edwards, H., Baker, B., Lee, T., Leike, J., Schulman, J., Sutskever, I., and Cobbe, K.
\newblock Let's verify step by step.
\newblock \emph{arXiv preprint arXiv:2305.20050}, 2023.

\bibitem[Lin et~al.(2022)Lin, Hilton, and Evans]{lin-etal-2022-truthfulqa}
Lin, S., Hilton, J., and Evans, O.
\newblock {T}ruthful{QA}: Measuring how models mimic human falsehoods.
\newblock In \emph{Proceedings of the 60th Annual Meeting of the Association for Computational Linguistics (Volume 1: Long Papers)}, pp.\  3214--3252, May 2022.

\bibitem[Liu et~al.(2022)Liu, Ash, Goel, Krishnamurthy, and Zhang]{liu2022transformers}
Liu, B., Ash, J.~T., Goel, S., Krishnamurthy, A., and Zhang, C.
\newblock Transformers learn shortcuts to automata.
\newblock \emph{arXiv preprint arXiv:2210.10749}, 2022.

\bibitem[Liu et~al.(2023)Liu, Yan, Zhang, An, Qiu, and Lin]{liu2023scaling}
Liu, X., Yan, H., Zhang, S., An, C., Qiu, X., and Lin, D.
\newblock Scaling laws of rope-based extrapolation.
\newblock \emph{arXiv preprint arXiv:2310.05209}, 2023.

\bibitem[Liu(2019)]{liu2019fine}
Liu, Y.
\newblock Fine-tune bert for extractive summarization.
\newblock \emph{arXiv preprint arXiv:1903.10318}, 2019.

\bibitem[Pang et~al.(2022)Pang, Parrish, Joshi, Nangia, Phang, Chen, Padmakumar, Ma, Thompson, He, and Bowman]{pang-etal-2022-quality}
Pang, R.~Y., Parrish, A., Joshi, N., Nangia, N., Phang, J., Chen, A., Padmakumar, V., Ma, J., Thompson, J., He, H., and Bowman, S.
\newblock {Q}u{ALITY}: Question answering with long input texts, yes!
\newblock In \emph{Proceedings of the 2022 Conference of the North American Chapter of the Association for Computational Linguistics: Human Language Technologies}, pp.\  5336--5358, July 2022.

\bibitem[Peng et~al.(2023)Peng, Quesnelle, Fan, and Shippole]{peng2023yarn}
Peng, B., Quesnelle, J., Fan, H., and Shippole, E.
\newblock Yarn: Efficient context window extension of large language models, 2023.

\bibitem[Press et~al.(2022)Press, Smith, and Lewis]{alibi}
Press, O., Smith, N., and Lewis, M.
\newblock Train short, test long: Attention with linear biases enables input length extrapolation.
\newblock In \emph{International Conference on Learning Representations (ICLR)}, 2022.

\bibitem[Rae et~al.(2020{\natexlab{a}})Rae, Potapenko, Jayakumar, Hillier, and Lillicrap]{Rae2020Compressive}
Rae, J.~W., Potapenko, A., Jayakumar, S.~M., Hillier, C., and Lillicrap, T.~P.
\newblock Compressive transformers for long-range sequence modelling.
\newblock In \emph{International Conference on Learning Representations (ICLR)}, 2020{\natexlab{a}}.

\bibitem[Rae et~al.(2020{\natexlab{b}})Rae, Potapenko, Jayakumar, Hillier, and Lillicrap]{pg19}
Rae, J.~W., Potapenko, A., Jayakumar, S.~M., Hillier, C., and Lillicrap, T.~P.
\newblock Compressive transformers for long-range sequence modelling.
\newblock In \emph{International Conference on Learning Representations}, 2020{\natexlab{b}}.

\bibitem[Raffel et~al.(2020)Raffel, Shazeer, Roberts, Lee, Narang, Matena, Zhou, Li, and Liu]{raffel2020exploring}
Raffel, C., Shazeer, N., Roberts, A., Lee, K., Narang, S., Matena, M., Zhou, Y., Li, W., and Liu, P.~J.
\newblock Exploring the limits of transfer learning with a unified text-to-text transformer.
\newblock \emph{The Journal of Machine Learning Research}, 21\penalty0 (1):\penalty0 5485--5551, 2020.

\bibitem[Ratner et~al.(2023)Ratner, Levine, Belinkov, Ram, Magar, Abend, Karpas, Shashua, Leyton-Brown, and Shoham]{ratner2023parallel}
Ratner, N., Levine, Y., Belinkov, Y., Ram, O., Magar, I., Abend, O., Karpas, E., Shashua, A., Leyton-Brown, K., and Shoham, Y.
\newblock Parallel context windows for large language models.
\newblock In \emph{Proceedings of the 61st Annual Meeting of the Association for Computational Linguistics (Volume 1: Long Papers)}, pp.\  6383--6402, 2023.

\bibitem[Roziere et~al.(2023)Roziere, Gehring, Gloeckle, Sootla, Gat, Tan, Adi, Liu, Remez, Rapin, et~al.]{roziere2023code}
Roziere, B., Gehring, J., Gloeckle, F., Sootla, S., Gat, I., Tan, X.~E., Adi, Y., Liu, J., Remez, T., Rapin, J., et~al.
\newblock Code llama: Open foundation models for code.
\newblock \emph{arXiv preprint arXiv:2308.12950}, 2023.

\bibitem[Ruoss et~al.(2023{\natexlab{a}})Ruoss, Del{\'e}tang, Genewein, Grau-Moya, Csord{\'a}s, Bennani, Legg, and Veness]{randompos}
Ruoss, A., Del{\'e}tang, G., Genewein, T., Grau-Moya, J., Csord{\'a}s, R., Bennani, M., Legg, S., and Veness, J.
\newblock Randomized positional encodings boost length generalization of transformers.
\newblock In \emph{Association for Computational Linguistics (ACL)}, July 2023{\natexlab{a}}.

\bibitem[Ruoss et~al.(2023{\natexlab{b}})Ruoss, Del{\'e}tang, Genewein, Grau-Moya, Csord{\'a}s, Bennani, Legg, and Veness]{ruoss-etal-2023-randomized}
Ruoss, A., Del{\'e}tang, G., Genewein, T., Grau-Moya, J., Csord{\'a}s, R., Bennani, M., Legg, S., and Veness, J.
\newblock Randomized positional encodings boost length generalization of transformers.
\newblock In \emph{Proceedings of the 61st Annual Meeting of the Association for Computational Linguistics (Volume 2: Short Papers)}, pp.\  1889--1903, July 2023{\natexlab{b}}.

\bibitem[Sakaguchi et~al.(2020)Sakaguchi, Le~Bras, Bhagavatula, and Choi]{sakaguchi2020winogrande}
Sakaguchi, K., Le~Bras, R., Bhagavatula, C., and Choi, Y.
\newblock Winogrande: An adversarial winograd schema challenge at scale.
\newblock In \emph{Proceedings of the AAAI Conference on Artificial Intelligence}, pp.\  8732--8740, 2020.

\bibitem[Shaham et~al.(2022)Shaham, Segal, Ivgi, Efrat, Yoran, Haviv, Gupta, Xiong, Geva, Berant, et~al.]{shaham2022scrolls}
Shaham, U., Segal, E., Ivgi, M., Efrat, A., Yoran, O., Haviv, A., Gupta, A., Xiong, W., Geva, M., Berant, J., et~al.
\newblock Scrolls: Standardized comparison over long language sequences.
\newblock \emph{arXiv preprint arXiv:2201.03533}, 2022.

\bibitem[Shaw et~al.(2018)Shaw, Uszkoreit, and Vaswani]{shaw2018self}
Shaw, P., Uszkoreit, J., and Vaswani, A.
\newblock Self-attention with relative position representations.
\newblock \emph{arXiv preprint arXiv:1803.02155}, 2018.

\bibitem[Su et~al.(2021)Su, Lu, Pan, Murtadha, Wen, and Liu]{rope}
Su, J., Lu, Y., Pan, S., Murtadha, A., Wen, B., and Liu, Y.
\newblock Roformer: Enhanced transformer with rotary position embedding, 2021.

\bibitem[Sun et~al.(2023)Sun, Dong, Patra, Ma, Huang, Benhaim, Chaudhary, Song, and Wei]{sun2022length}
Sun, Y., Dong, L., Patra, B., Ma, S., Huang, S., Benhaim, A., Chaudhary, V., Song, X., and Wei, F.
\newblock A length-extrapolatable transformer.
\newblock In \emph{Proceedings of the 61st Annual Meeting of the Association for Computational Linguistics (Volume 1: Long Papers)}. Association for Computational Linguistics, July 2023.

\bibitem[Touvron et~al.(2023)Touvron, Martin, Stone, Albert, Almahairi, Babaei, Bashlykov, Batra, Bhargava, Bhosale, Bikel, Blecher, Ferrer, Chen, Cucurull, Esiobu, Fernandes, Fu, Fu, Fuller, Gao, Goswami, Goyal, Hartshorn, Hosseini, Hou, Inan, Kardas, Kerkez, Khabsa, Kloumann, Korenev, Koura, Lachaux, Lavril, Lee, Liskovich, Lu, Mao, Martinet, Mihaylov, Mishra, Molybog, Nie, Poulton, Reizenstein, Rungta, Saladi, Schelten, Silva, Smith, Subramanian, Tan, Tang, Taylor, Williams, Kuan, Xu, Yan, Zarov, Zhang, Fan, Kambadur, Narang, Rodriguez, Stojnic, Edunov, and Scialom]{touvron2023llama}
Touvron, H., Martin, L., Stone, K., Albert, P., Almahairi, A., Babaei, Y., Bashlykov, N., Batra, S., Bhargava, P., Bhosale, S., Bikel, D., Blecher, L., Ferrer, C.~C., Chen, M., Cucurull, G., Esiobu, D., Fernandes, J., Fu, J., Fu, W., Fuller, B., Gao, C., Goswami, V., Goyal, N., Hartshorn, A., Hosseini, S., Hou, R., Inan, H., Kardas, M., Kerkez, V., Khabsa, M., Kloumann, I., Korenev, A., Koura, P.~S., Lachaux, M.-A., Lavril, T., Lee, J., Liskovich, D., Lu, Y., Mao, Y., Martinet, X., Mihaylov, T., Mishra, P., Molybog, I., Nie, Y., Poulton, A., Reizenstein, J., Rungta, R., Saladi, K., Schelten, A., Silva, R., Smith, E.~M., Subramanian, R., Tan, X.~E., Tang, B., Taylor, R., Williams, A., Kuan, J.~X., Xu, P., Yan, Z., Zarov, I., Zhang, Y., Fan, A., Kambadur, M., Narang, S., Rodriguez, A., Stojnic, R., Edunov, S., and Scialom, T.
\newblock Llama 2: Open foundation and fine-tuned chat models, 2023.

\bibitem[Vaswani et~al.(2017)Vaswani, Shazeer, Parmar, Uszkoreit, Jones, Gomez, Kaiser, and Polosukhin]{vaswani2017attention}
Vaswani, A., Shazeer, N., Parmar, N., Uszkoreit, J., Jones, L., Gomez, A.~N., Kaiser, {\L}., and Polosukhin, I.
\newblock Attention is all you need.
\newblock \emph{Advances in Neural Information Processing Systems (NeurIPS)}, 30, 2017.

\bibitem[Wei et~al.(2022)Wei, Wang, Schuurmans, Bosma, brian ichter, Xia, Chi, Le, and Zhou]{wei2022chain}
Wei, J., Wang, X., Schuurmans, D., Bosma, M., brian ichter, Xia, F., Chi, E.~H., Le, Q.~V., and Zhou, D.
\newblock Chain of thought prompting elicits reasoning in large language models.
\newblock In Oh, A.~H., Agarwal, A., Belgrave, D., and Cho, K. (eds.), \emph{Advances in Neural Information Processing Systems}, 2022.

\bibitem[Xiao et~al.(2023)Xiao, Tian, Chen, Han, and Lewis]{xiao2023efficient}
Xiao, G., Tian, Y., Chen, B., Han, S., and Lewis, M.
\newblock Efficient streaming language models with attention sinks, 2023.

\bibitem[Zellers et~al.(2019)Zellers, Holtzman, Bisk, Farhadi, and Choi]{zellers2019hellaswag}
Zellers, R., Holtzman, A., Bisk, Y., Farhadi, A., and Choi, Y.
\newblock Hellaswag: Can a machine really finish your sentence?
\newblock In \emph{Proceedings of the 57th Annual Meeting of the Association for Computational Linguistics}, 2019.

\bibitem[Zhong et~al.(2021)Zhong, Yin, Yu, Zaidi, Mutuma, Jha, Awadallah, Celikyilmaz, Liu, Qiu, and Radev]{zhong2021qmsum}
Zhong, M., Yin, D., Yu, T., Zaidi, A., Mutuma, M., Jha, R., Awadallah, A.~H., Celikyilmaz, A., Liu, Y., Qiu, X., and Radev, D.
\newblock {QMS}um: A new benchmark for query-based multi-domain meeting summarization.
\newblock In \emph{Proceedings of the 2021 Conference of the North American Chapter of the Association for Computational Linguistics: Human Language Technologies}, pp.\  5905--5921, June 2021.

\bibitem[Zhu et~al.(2023)Zhu, Yang, Wang, Song, Wu, Wei, and Li]{zhu2023pose}
Zhu, D., Yang, N., Wang, L., Song, Y., Wu, W., Wei, F., and Li, S.
\newblock Pose: Efficient context window extension of llms via positional skip-wise training.
\newblock \emph{arXiv preprint arXiv:2309.10400}, 2023.

\end{thebibliography}
\bibliographystyle{icml2024}

\newpage
\appendix
\onecolumn

\section{Proofs}
\label{sec:Proof}

\subsection{Technical Lemmas}
In this subsection, we present some technical lemmas that show how the MLP and Transformer can perform basic operations. We show that the MLP with GeLU activation can perform scalar multiplication, selection, and Boolean matrix multiplication; and that the attention layer can perform the COPY operation.

\begin{lemma}[From \citet{feng2023towards}]
\label{lemma:MLP_multip}
    For any $\epsilon > 0$ and $M > 0$, there exist a two-layer MLP $f:\mathbb R^2\to \mathbb R$ with $\mathrm{GeLU}$ activation and parameters with $\ell_{\infty}$ norm upper bounded by $O(\mathrm{poly}(M,1/\eps))$ such that $|f(a, b) - ab| \leq \epsilon$ holds for all $a, b \in [-M, M]$.
\end{lemma}

\begin{lemma}[From \citet{feng2023towards}]
\label{lemma:MLP_relu}
    Let $\vg:\mathbb R^{d_1}\to \mathbb R^{d_2}$ be a two-layer MLP with $\mathrm{ReLU}$ activation, and all parameter values are upper bounded by $M$. For any $\epsilon>0$, there exists a two-layer MLP $\vf$ of the same size with $\mathrm{GeLU}$ activation and parameters upper bounded by $O(\mathrm{poly}(M, 1/\epsilon))$ in the $\ell_{\infty}$ norm, such that for all $\vx\in\mathbb R^{d_1}$, we have $\|\vf(\vx)-\vg(\vx)\|_\infty\leq \epsilon$.
\end{lemma}

\begin{lemma}[From \citet{feng2023towards}]
\label{lemma:MLP_select}
    Define the selection function $\vg:\mathbb R^{d}\times \mathbb R^{d}\times \mathbb R\to \mathbb R^{d}$ as follows:
    \begin{equation}
        \vg(\vx,\vy,t)=\left\{\begin{array}{cc}
            \vx & \text{if }t\ge 0, \\
            \vy & \text{if }t< 0.
        \end{array}\right.
    \end{equation}
    Let $\vf:\mathbb R^{d}\times \mathbb R^{d}\times \mathbb R\to \mathbb R^{d}$ be a two-layer MLP with $\mathrm{GeLU}$ activation. Then, for any $\epsilon>0$, $\alpha>0$, and $M>0$, there exist MLP parameters with $\ell_{\infty}$ norm bounded by $O(\mathrm{poly}(M,1/\alpha, 1/\epsilon))$, such that for all $\vx\in[-M,M]^{d}$, $\vy\in [-M,M]^{d}$, and $t\in [-\infty,-\alpha]\cup[\alpha,+\infty]$, we have $\|\vf(\vx,\vy,t)-\vg(\vx,\vy,t)\|_\infty\leq \epsilon$.
\end{lemma}

\begin{lemma}
    \label{lemma:MLP_matrix}
    Let $\vf:\mathbb R^{d_1\times d_2}\times \mathbb R^{d_2\times d_3}\times \to \mathbb R^{d_1\times d_3}$ be a two-layer MLP with $\mathrm{GeLU}$ activation, and given a Boolean matrix $\mB$, let $\ve_{\mB}$ be the vector by flattening the matrix $\mB$. Then, for any $\epsilon>0$, $\alpha>0$, and $M>0$, there exist MLP parameters with $\ell_{\infty}$ norm bounded by $O(\mathrm{poly}(1/\epsilon))$, such that for all $\mA\in\{0,1\}^{d_1\times d_2}$ and $\mB\in \{0,1\}^{d_2\times d_3}$, we have $\|\vf(\ve_\mA,\ve_\mB)-\ve_{\mA\cdot \mB}\|_\infty\leq \epsilon$.
\end{lemma}

The proof of \cref{lemma:MLP_multip,lemma:MLP_relu,lemma:MLP_select} can be found in the appendix of \citet{feng2023towards}, and we will give the proof of \cref{lemma:MLP_matrix}.

\begin{proof}[Proof of \cref{lemma:MLP_matrix}]
    Given two Boolean matrices $\mA\in\{0,1\}^{d_1\times d_2}$ and $\mB\in \{0,1\}^{d_2\times d_3}$. We can represent the output $\mA\cdot\mB$ by the following formula:
    \begin{align*}
        &(\mA\cdot\mB)_{i,j}=\bigvee_{k\in[d_2]} (\mA_{i,k}\wedge\mB_{k,j})\\
        =&\text{ReLU}\Big(\sum_{k\in[d_2]}\text{ReLU}(\mA_{i,k}+\mB_{k,j}-1)\Big)-\text{ReLU}\Big(\sum_{k\in[d_2]}\text{ReLU}(\mA_{i,k}+\mB_{k,j}-1)-1\Big) 
    \end{align*}
    Therefore, we can implement the Boolean matrix multiplication by the MLP with ReLU activation, and according to \cref{lemma:MLP_relu}, the MLP with GeLU activation can perform the Boolean matrix multiplication.
\end{proof}

Then we introduce a basic operation that can be implemented by the attention layer. And following it, we give a special form of this operation used in the proof of our main theorems.

Let $n$ be an integer and $\ve_1, \ve_2, \cdots, \ve_n$ be a sequence of vectors, whose $\ell_\infty$ norm is bounded by a large constant $M$. Let $\mK,\mQ\in\mathbb R^{d'\times d}$ be any matrices, and let $0<\rho <M$ be any real numbers. Denote $\vq_i=\mQ\vx_i$ and $\vk_j=\mK\vx_j$. The output of the COPY operation is a sequence of vectors $\vu_1,\cdots,\vu_n$ with $\vu_i=\ve_{\mathrm{pos}(i)}$, where $\mathrm{pos}(i)=\operatorname{argmax}_{j\in [i]}{\vq_i\cdot\vk_j}$. Moreover, we assume that the matrices $\mQ,\mK$ and scalars $ \delta$ satisfy that for all considered sequences $\ve_1, \ve_2, \cdots, \ve_n$, we have for any $i$ and $j\in [n]\backslash \{\text{pos}(i)\}$, either $\vq_i\cdot \vk_{\text{pos}(i)}-\vq_i\cdot \vk_j\ge \delta$. This assumption guarantees that there are sufficient gaps between the attended position and other positions. Then we prove that the attention layer can implement the COPY operation.

\begin{lemma}[From \citet{feng2023towards}]
\label{lemma:TM_copy}
      For any $\epsilon > 0$, there exists an attention layer with embedding size $O(d)$ and one causal attention head that can approximate the COPY operation defined above. Formally, for any considered sequence of vectors $\ve_1, \ve_2, \dots, \ve_n$, denote the corresponding attention output as $\vo_1, \vo_2, \dots, \vo_n$. Then, we have $\|\vo_i-\vu_i\|_{\infty}\le\epsilon$ for all $i\in [n]$. Moreover, the $\ell_\infty$ norm of attention parameters is bounded by $O(\mathrm{poly}(M,1/\delta,\log(n),\log(1/\epsilon)))$.
\end{lemma}

The proof of this lemma can be found in \citet{feng2023towards}. Based on the \cref{lemma:TM_copy,lemma:MLP_multip}, given the token index $i$ as the absolute PE, the attention layer can copy the embedding at the specific position by the following construction. Given a specific position $i$, we can construct the query $\vq=(1,i^2,i)$ and the key of the $j$-th token $\vk_j=(-j^2,-1,2j)$ by MLP, according to \cref{lemma:MLP_multip}. Then $\vq\cdot \vk_j=-(i-j)^2$ gets the maximum when $i=j$, and we can concentrate the attention on the $i$-th token and copy its embedding by \cref{lemma:TM_copy}.

\subsection{Proofs of Main Theorems}
\label{sec:proof_main_theorem}
In this subsection, we will prove \cref{Thm:BiPE,Thm:classic_PE}, for ease of reading we restate the theorems here and then give proof.

\subsubsection{Proof of Theorem \ref{Thm:classic_PE}}
\begin{theorem}[Lower Bound for Transformer with APE to Represent NFA]
\label{Thm:classic_PE:app}
    For any size of state set, there exists an NFA $\gN=(Q,\Sigma,\delta,q_0,F)$ such that a Transformer with APE needs at least $O(|Q|^2)$ embedding size to represent the NFA.
\end{theorem}

We prove this theorem in the log-precision setting, which is a realistic and practical setting for transformers. In this setting, each value in the transformer is encoded by $O(\log N)$ bits, where $N$ is the input length. This corresponds to the practical scenario where the transformer handles input sequences of up to a few thousand tokens, using $16$ or $32$ bits floating-point numbers. Theoretically, the log-precision number can approximate any real number of magnitude $O(\mathrm{poly}(N))$ with an error of $O(\mathrm{poly}(1/N))$. Each neuron in the transformer can store only $O(\log(n))$-bits information and thus cannot retain the full information of the entire input sequence, which is reasonable and aligned with practical scenarios.

\begin{proof}[Proof of \cref{Thm:classic_PE:app}]
    Given the state set $Q$, we can construct a NFA $\gN=(Q,\Sigma,\delta,q_0,F)$ as follows:
    \begin{itemize}
        \item The state set $Q$ is given, and assuming $Q=\{q_1,q_2,\cdots,q_n\}$.
        \item $\Sigma$ is the set of all maps from the state set $Q$ to its power set, i.e. $\Sigma=\{f:Q\rightarrow\gP(Q)\}$.
        \item Given a state $q\in Q$ and a symbol $f\in \Sigma$, $\delta(q,f)=f(q)$.
        \item $q_0=q_1$.
        \item $F=\{q_n\}$.
    \end{itemize}
    To prove that the embedding dimension of the transformer to represent $\gN$ is at least $O(n^2)$, we use the following argument. The size of the alphabet is $2^{n^2}$, so we need at least $O(n^2)$ bits to embed each input symbol. Suppose the embedding dimension is $o(n^2)$. Then we can find two input sequences of constant size that have the same embeddings in the transformer but different outcomes for the NFA. Since the length of the input sequence is constant, the transformer uses $o(n^2)$ bits to represent the embedding for each token. Therefore, there exist two input tokens $f$ and $f^\prime$ that have the same embedding. Let $q_i\in f(q_j)$ and $q_i\notin f^\prime(q_j)$. Then we can construct two input sequences $\mS=[f_1,f,f_2]$ and $\mS^\prime=[f_1,f^\prime,f_2]$, where $f_1(q_0)=\{q_j\}$, $f_2(q_i)=\{q_n\}$, and $f_2(q)=\emptyset$ for $q\neq q_i$. The embeddings of these two sequences in the transformer are the same, but one is accepted by $\gN$ while the other is not. Hence, the transformer with embedding size $o(n^2)$ cannot represent the NFA $\gN$. A transformer with APE needs at least $O(n^2)$ embedding size to represent the NFA.
\end{proof}

\subsubsection{Proof of Theorem \ref{Thm:BiPE}}
\begin{theorem}[Upper Bound for Transformer with BiPE to Represent Bi-NFA]
\label{Thm:BiPE:app}
    \looseness=-1For any Bi-NFA $\gN=(\gQ,\Sigma,\delta,q_0,F)$, $\gQ=\{Q_1,Q_2,\cdots,Q_k\}$ there exists a Transformer with BiPE and $O(k^2+\sum_{i\in[k]} |Q_i|^2)$ embedding size can represent the Bi-NFA.
\end{theorem}

In this proof, we use the log-precision transformer with the GeLU activation function and $O(\log N)$ layers, where $N$ is the input length. This choice of layers is crucial for the transformer to represent automata, as a constant-layer transformer would require a super-polynomial embedding size (in the input length) to do so \cite{liu2022transformers}. Without loss of generality, we focus on the Bi-PE model with T5-relative PE as the inter-segment positional encoding and APE as the intra-segment positional encoding. Moreover, our proof can be easily extended to other variants of Bi-PE, such as those with RoPE or AliBi as the inter-segment positional encoding. 

\paragraph{T5-relative PE.} We use the T5-relative PE method to compute the attention logits before applying the softmax function. The attention logits are given by the following equation:
\begin{equation*}
    \mA_\text{RPE}(\mX)=\mX\mW_Q(\mX\mW_K)^\top+\mB
\end{equation*}
where $\mB_{i,j}=r_{\min(i-j,K)}$, $K$ is a hyper-parameter, and $\{r_{i}\}_{i=0}^K$ are learnable scalars that represent the relative position embeddings.

\begin{proof}[Proof Sketch]
    In this proof, we construct a transformer containing two modules to represent the Bi-NFA. The first module computes the state transitions in the segment, and the second module computes the state transitions between the segments. Each module contains $O(\log N)$ attention layers and implements a classic divide-and-conquer algorithm. 
\end{proof}

\begin{proof}[Proof of \cref{Thm:BiPE:app}]
    Given a Bi-NFA $\gN=(\gQ,\Sigma,\delta,q_0,F)$, we will first introduce some notations and the basic idea behind our construction. 

    \paragraph{The State Transition in Each Segment.} Given $\gQ=\{Q_1,\cdots,Q_k\}$, we use $q_{i,j}$ for the $j$-th state in $Q_i$ and define $Q=\bigcup_{i\in[k]}Q_i$, $Q^*=\{q_i^*\}$. Without loss of generality, we assume that $q_{i,1}$ is the start state $q_i^*$ in $Q_i$ and $q_0=q_1^*$. Given an input symbol $w$, we view it as a map $f_w$ from $Q$ to $\gP(Q)$ such that $f_w(q)=\delta(w,q)$, and an input string $\mS=[w_1,w_2,\cdots, w_n]$ as the composition of $f_s=f_{w_1}\cdot f_{w_2}\cdots f_{w_n}$. Denoting the segment separator as $w^*\in\Sigma$, for any input symbol $w\neq w^*$, we use a tuple of boolean matrices $\tM(w)=\big(\mM_1(w),\mM_2(w),\cdots,\mM_n(w)\big)$ to represent it, such that $\tM_{i,j,k}(w)=\big(\mM_i(w)\big)_{j,k}=\mathbb{I}[q_{i,j}\in\delta(w,q_{i,k})]$. We define the multiplication of two tuples of matrices as $\tM(w)\cdot\tM(w^\prime)=\big(\mM_1(w)\cdot\mM_1(w^\prime),\cdots,\mM_n(w)\cdot\mM_n(w^\prime)\big)$, which is the composition of $f_w\cdot f_{w^\prime}$. In the transformer, we flatten $\tM(w)$ to a vector $\vm(w)$. We use the MLP to implement the multiplication of two tuples of matrices according to \cref{lemma:MLP_matrix}. Moreover, a string without a segment separator can be viewed as the multiplication of these tuples of matrices. Given a string without segment separator $S=[w_1,w_2,\cdots, w_n]$, we compute $\tM(S)=\prod_{i\in[n]}\tM(w_i)$ and denote the vector flattened from $\tM(S)$ as $\vm(S)$. Given the start state $q_i^*$, we get the state set of the Bi-NFA $\{q_{i,j}|\tM_{i,1,j}(s)=1\}$ after taking in the string $S$. Moreover, we use the notation $w_{i,j}:w_{i,k}$ and $\tM(w_{i,j}:w_{i,k})$ to represent the substring from $w_{i,j}$ to $w_{i,k}$ and its state transition matrix tuple. We can use the classic divide-and-conquer algorithm to compute $\tM(s)=\prod_{i\in[n]}\tM(w_i)$, and the first module of the transformer we construct implements the algorithm.

    \paragraph{The State Transition between Segments.}
    A segment $S=[w_1,w_2,\cdots, w_n,w^*]$ can be viewed as a map from $\{q_1^*,q_2^*,\cdots,q_k^*\}$ to $\gP(\{q_1^*,q_2^*,\cdots,q_k^*\})$. For the segment separator $w^*$, we can also view it as a tuple of matrices $\tM(w^*)=\big(\mM_1(w^*),\mM_2(w^*),\cdots,\mM_k(w^*)\big)$, where $\tM_{i,j,k}(w^*)=\mathbb{I}[q^*_k\in\delta(w^*,q_{i,j})]$. Therefore, we can compute the state transition matrix $\mA(S)$ of the segment $S$ such that $\mA_{i,j}(S)=\tM_{i,1,j}(S)$. We have $\mA_{i,j}(S)=\mathbb{I}[q_j^*\in \delta(S,q_i^*)]$. Given a sequence of segments $\mS=S_1\oplus S_2 \oplus\cdots\oplus S_n$, we can compute $\mA(\mS)=\Pi_{i\in[n]}\mA(S_i)$. Then given the start state $q_0$, we can get the final state set $Q(\mS)$ after the Bi-NFA taking in the input $\mS$ such that $Q(\mS)=\{q_i^*|\mA_{1,i}(\mS)=1\}$. Then the Bi-NFA accepts $\mS$ if and only if $Q(\mS)\cap F\neq \emptyset$, and this condition judgment can be formulated as the product of two Boolean vectors and therefore, can be implemented by MLP. For convenience and clarity in presenting our proof, in the representation of the transformer, we flatten the matrices $\mA(\mS)$ and $\mA(S)$ to vectors and we denote them as $\va(\mS)$ and $\va(S)$, respectively. We use the notation $S_i:S_j$ and $\mA(S_i:S_j)$ to represent the substring from $S_i$ to $S_j$ and its state transition matrix.

    \paragraph{Token Embeddings.} 
    Given a input sequence $\mS=S_1\oplus S_2 \oplus\cdots\oplus S_n$ and $S_i=[w_{i,1},w_{i,2},\cdots w_{i,l_i-1},w^*_i]$, assuming the length of the sequence is upper boumded by $N$. As a standard way in NLP, we add a \texttt{<SOS>} at the beginning of the string. For each token $w_{i,j}\neq w^*,$ and $w_{i,j}\neq \texttt{<SOS>}$, after combining the absolute positional embedding, the embedding at the beginning is $\vx^0_{i,j}=\big(\vm(w_{i,j}),1,0,0,i\big)$, and the embeddings for $w^*$ and \texttt{<SOS>} is $(\mathbf{0},0,1,0,l_i)$ and $(\mathbf{0},0,0,1,1)$, respectively. The embedding size at begining is $O(\sum_{i\in[k]} |Q_i|^2)$. 

    \paragraph{Module \uppercase\expandafter{\romannumeral1}.}
    The first module contains $(\lceil\log(N)\rceil+1)$ layers, and in this module, the token only attends to the tokens in the same segment. Therefore, the T5-relative PE for these layers is $r_i=-\infty$ for $i\neq 0$ and $r_0=0$. At the layer $l$, the input embeddings of token $w_{i,j}$ is $\vx^{1,l}_{i,j}=\big(\vm(w_{i,\max(1,j-2^l)}:w_{i,j}),1,0,0,i\big)$, where $w_{i,j_1}:w_{i,j_2}=[w_{i,j_1},w_{i,j_1+1},\cdots, w_{i,j_2}]$. The first module completes the following tasks:
    \begin{itemize}
        \item Copy the embedding of token $w_{i,j-2^l}$, note that when $j-2^l<1$, the embedding copied is meaningless.
        \item Calculate the multiplication of two tuples of transition matrices defined in the previous paragraph.
        \item Select the embedding, when $j-2^l<1$, the output embedding is the same as the input, and when $j-2^l\geq 1$, the output is the outcome of the multiplication.
    \end{itemize}
    Therefore, the output embeddings of the token $w_{i,j}$ at the layer $l$ is $\vx^{1,l+1}_{i,j}=\big(\vm(w_{i,\max(1,j-2^(l+1))}:w_{i,j}),1,0,0,i\big)$. According to \cref{lemma:MLP_matrix,lemma:MLP_select,lemma:TM_copy} we can implement the COPY operation by the attention layer, and implement the multiplication of two tuples of matrices, and the selection operation by the MLP. After $\lceil\log(N)\rceil$ layers, the output embedding of $w_{i,j}$ is $\vx^{1,\lceil\log(N)\rceil}_{i,j}=\big(\vm(w_{i,1}:w_{i,j}),1,0,0,i\big)$. At the final layer, the token $w^*$ copies the embedding of the previous token $w_{i,l_i-1}$, and uses the MLP to compute the transition matrix of this segment. The final output of this block for the token $w^*_i$ is $\big(\va(S_i),0,1,0,l_i\big)$, for the token \texttt{<SOS>} is $(\mathbf{0},0,0,1,1)$, and for the token $w_{i,j}$ is $(\mathbf{0},1,0,0,j)$. The embedding size of the first module is $O(\sum_{i\in[k]} |Q_i|^2)$.

    \paragraph{Module \uppercase\expandafter{\romannumeral2}.}
    The second module contains $\lceil\log(N)\rceil$ layers, and we only need to concentrate our attention on the embedding of the last token of each segment. Similar to the previous block, at the layer $l$, the input embeddings of token $w^*_{i}$ is $\big(\va(S_{\max(1,i-2^l)}:S_{i}),0,1,0,l_i\big)$, where $S_{i}:S_{j}=S_i\oplus S_{i+1}\oplus\cdots\oplus S_j$.  The second module completes the following tasks:
    \begin{itemize}
        \item Copy the embedding of last token of segment $S_{i-2^l}$, note that when $i-2^l<1$, the embedding copied is meaningless.
        \item Calculate the multiplication of two transition matrices defined in the previous paragraph.
        \item Select the embedding: when $i-2^l<1$, the output embedding is the same as the input, and when $i-2^l\geq 1$, the output is the outcome of the multiplication.
    \end{itemize}
    \looseness=-1Therefore, the output embeddings of the token $w^*_{i}$ is $\big(\va(S_{\max(1,i-2^{(l+1)})}:S_{i}),0,1,0,l_i\big)$. we design the relative positional embedding as $r_i=-\infty$ for $i\neq 2^l$ and $r_{2^l}=0$. Note that, when $i-2^l<1$, the token will give uniform attention to the last token of each segment previous to it, therefore, we can use the embedding of \texttt{<SOS>} to detect this case. When the value indicates \texttt{<SOS>} is greater than $\frac{1}{N}$, we have $i-2^l<1$ and maintain the embeddings. According to \cref{lemma:MLP_matrix,lemma:MLP_select,lemma:TM_copy} we can implement the copy operation by the attention layer, and implement the multiplication of two matrices and the selection operation by the MLP. Then we can get the final state set from the embedding of the last token of the input sequence and we can use an MLP to compute the outcome to determine accepting the input or not. The embedding size of the second module is $O(k^2)$.

    Therefore, we construct a transformer with BiPE, and $O(k^2+\sum_{i\in[k]} |Q_i|^2)$ embedding size can represent the Bi-NFA.
\end{proof} 

\section{More Related Works}

\subsection{Previous Bi-level Positional Encoding Approaches.} \label{appx:more-related-works}

The original BERT~\cite{bert} model also includes two encodings for representing positions, but its approach differs significantly from our BiPE. Primarily, BERT only needs to specify two segments using absolute encoding, tailored for the next sentence prediction task, not for length extrapolation. Furthermore, BERT treats a sequence as a flat array of tokens and defines the segments in an arbitrary way, ignoring the intrinsic segmentation of language data. \citet{liu2019fine} further extends BERT for summarization tasks. They modify the BERT configuration by encoding whether the segment index is odd or even and encoding the absolute token position in the whole sequence. It is noteworthy that the first one cannot carry enough segment-level positional signals, and the latter one faces the same problem as BERT and cannot extrapolate to longer contexts. Another similar work is Segatron~\cite{bai2021segatron}. It introduces paragraph and sentence segmentation to the relative position encoding, which resembles the inter-segment encoding in BiPE. However, the absolute token position is still calculated in the whole sequence, which still incurs length extrapolation issues. In contrast, in our work, the intra-segment encoding only identifies the location of each token within the segment, and the inter-segment encoding specifies the segment indexes. By properly using different kinds of positional encodings, our BiPE can be used in longer sequences.
\section{Visualization of Distribution}\label{app_exp:add_distribution}

\begin{figure*}[h]
    \includegraphics[width=0.995\linewidth]{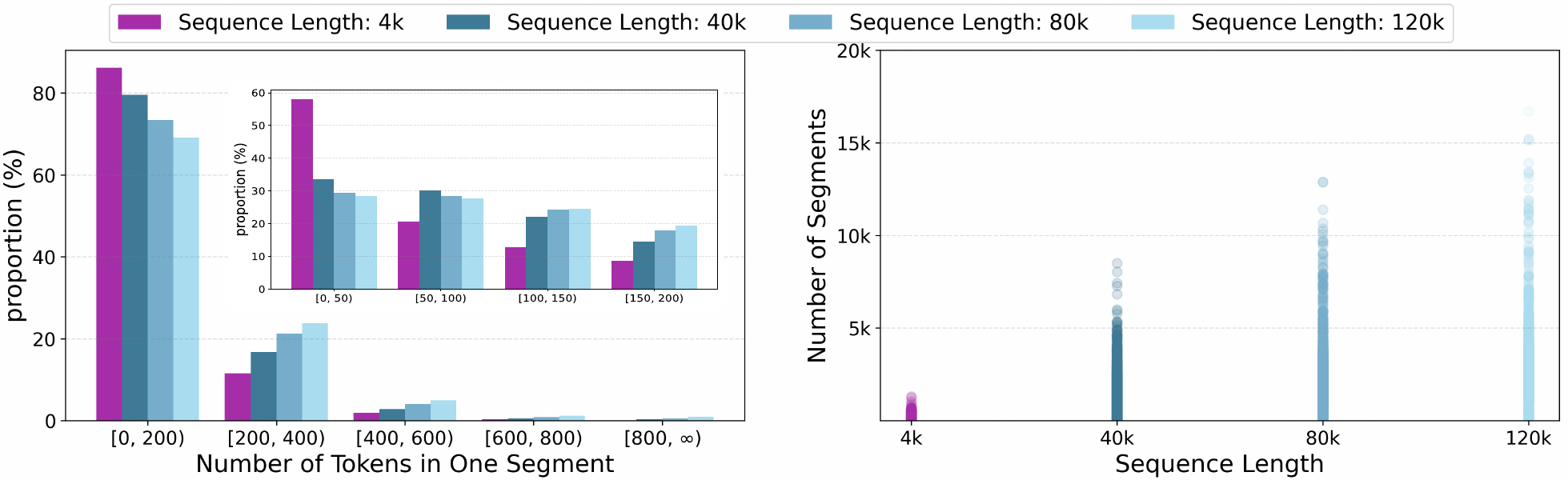}
    \caption{\textbf{Left:} The distribution of the token number in one segment with different sequence lengths. \textbf{Right:} The distribution of the number of segments with different sequence lengths.  We use the tokenizer of Llama 2~\cite{touvron2023llama} for tokenization on ArXiv. Full stop``\texttt{.}'' and newline ``\texttt{\textbackslash n}'' are used for segmentation.
    }
    \label{fig:arxiv_token_distribution}
\end{figure*}

\begin{figure*}[h]
    \includegraphics[width=0.995\linewidth]{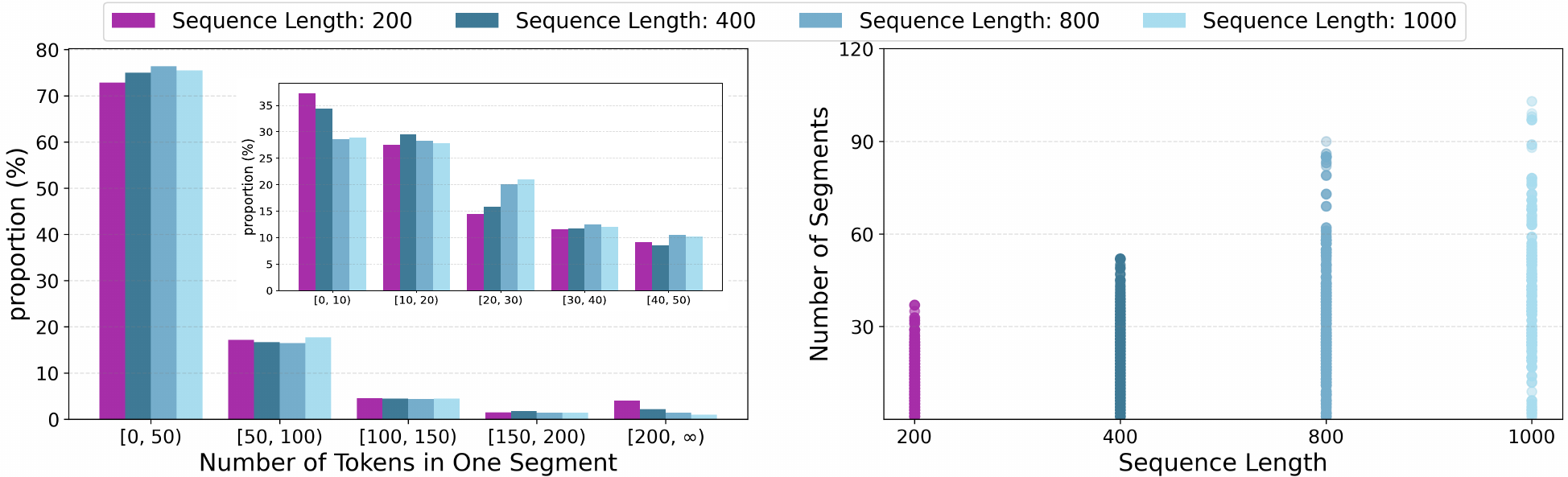}
    \caption{\textbf{Left:} The distribution of the token number in one segment with different sequence lengths. \textbf{Right:} The distribution of the number of segments with different sequence lengths.  We use the tokenizer of Llama 2~\cite{touvron2023llama} for tokenization on PRM800K~\cite{lightman2023lets}. Full stop``\texttt{.}'' and newline ``\texttt{\textbackslash n}'' are used for segmentation.
    }
    \label{fig:prm800k_token_distribution}
\end{figure*}
\newpage
\section{Experimental Details}\label{app_exp:add_exp}

\subsection{Capacity Experiments}\label{app:cot_math}

\begin{table}[!ht]
\centering
\begin{minipage}{.5\textwidth}
\caption{Model configurations for capacity experiments.}
\label{tab:app_math_cot_model}
\centering
\begin{tabular}{ccc}
\toprule
    Layers & & $3$\\
    Attention heads & & $4$\\
    Head dimensions & & $\{12, 16, 64\}$\\
    Hidden dimensions & & $\{48, 64, 256\}$\\
    FFN dimensions & & $\{192, 256, 1024\}$\\
    Model parameters & & \{$87$K, $153$K, $2.4$M\}\\
\bottomrule
\end{tabular}
\end{minipage}%
\hfill
\begin{minipage}{.5\textwidth}
\caption{Training recipes for capacity experiments.}
\label{tab:app_cot_math_train}
\centering
\begin{tabular}{ccc}
\toprule
    Batch size & & $512$ \\
    Epochs & & $100$  \\
    Dropout & & $0.1$  \\
    Weight decay & & $0.01$  \\
    Optimizer & & AdamW \\
    Learning rate & & $1\mathrm{e}-4$   \\
\bottomrule
\end{tabular}
\end{minipage}
\end{table}

In this experiment, we use the Arithmetic task~\cite{feng2023towards} to empirically verify the parameter efficiency brought by our BiPE method. Given an arithmetical expression consisting of numbers, basic operations ($+,-,\times,\div,=$) and brackets, e.g., $(1+2)\times(3+5)=$, this task requires language models to calculate and generate the correct result, e.g., $24$. Following~\citet{feng2023towards}, we train all models using Chain-of-Thought demonstrations, e.g., for the input sequence $(7+8)\div(5+2\times 7-2\times 8)$, the output sequence is $15\div(5+2\times 7-2\times 8)=15\div(5+14-2\times 8)=15\div(19-2\times 8)=15\div(19-16)=15\div 3=5$. The evaluation metric is the accuracy of the final answer. We refer interested readers to Appendix H in \citet{feng2023towards} for additional details.

\paragraph{Model configurations.} In this experiment, we train decoder-only Transformer-based language models with different positional encoding techniques while keeping all the other configurations the same. For Sinusoidal PE, we follow \citet{vaswani2017attention} to set the hyperparameters in sine and cosine functions. For RoPE and XPOS, we follow \citet{rope,sun2022length} to set the hyperparameters in the rotary matrix respectively. For ALiBi, we follow \citet{alibi} to set the slope values in each attention head. For the intra segment encoding of our BiPE, we use the learnable absolute positional encoding. For the inter segment encoding of our BiPE-RoPE, the hyperparameters are kept the same as~\cite{rope}. For the inter segment encoding of our BiPE-ALiBi, the slope values are set to 96 times of the original ALiBi's setting. Other model configurations are provided in Table \ref{tab:app_math_cot_model}.

\paragraph{Training recipes.} The next token prediction objective~\cite{brown2020language} is adopted for language model training. The number of operators in the arithmetic dataset is set to 6, which yields a total sequence length of 223 for Chain-of-Thought demonstrations. The training recipes are provided in Table~\ref{tab:app_cot_math_train}. All models are trained on 4 NVIDIA A100 GPUs.

\subsection{Length Extrapolation Experiments}\label{app:extra_exp}
\begin{table}[!ht]
\centering
\begin{minipage}{.5\textwidth}
\caption{Model configurations for length extrapolation experiments.}\label{tab:app_lm_model}
\centering
\begin{tabular}{ccc}
\toprule
    Layers & & $12$\\
    Attention heads & & $12$\\
    Head dimensions & & $64$\\
    Hidden dimensions & & $768$\\
    FFN dimensions & & $3072$\\
    Model parameters & & $155$M\\ 
\bottomrule
\end{tabular}
\end{minipage}%
\hfill
\begin{minipage}{.45\textwidth}
\caption{Training recipes for length extrapolation experiments.}\label{tab:app_lm_train}
\centering
\begin{tabular}{ccc}
\toprule
    Batch size & & $256$ \\
    Total training steps & & $500$k  \\
    Dropout & & $0.0$  \\
    Weight decay & & $0.01$  \\
    Optimizer & & AdamW \\
    Learning rate & & $1\mathrm{e}-4$   \\
\bottomrule
\end{tabular}
\end{minipage}
\end{table}

\begin{figure*}[ht]
    \begin{center}
    \includegraphics[width=1.0\linewidth]{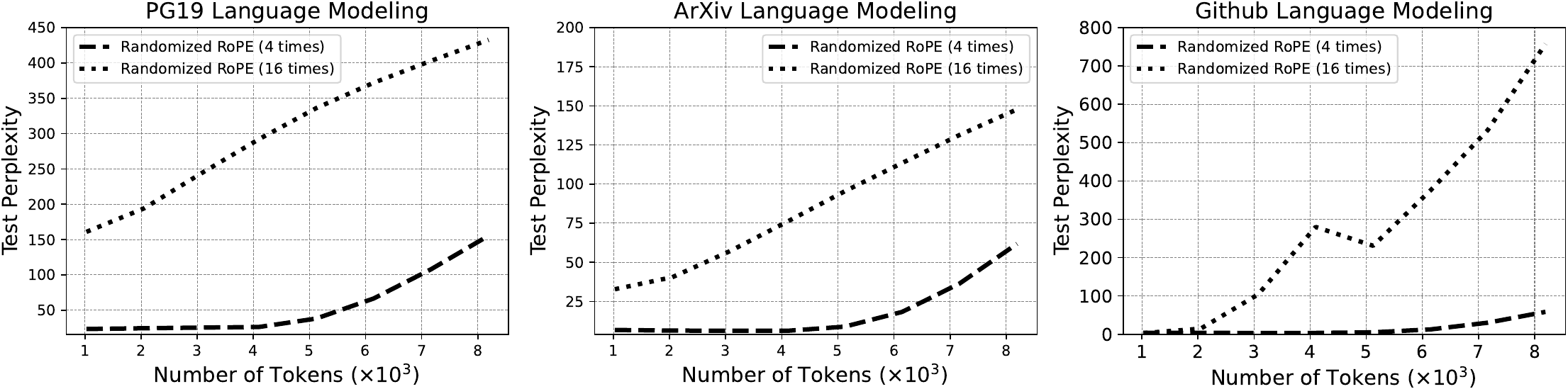}
    \end{center}
    \vspace{-0.5em}
    \caption{Language modeling perplexity with varying evaluation sequence lengths for Randomized RoPE trained on the Pile dataset with different times of the training length for extended positions.}
    \label{fig:app_randomized_rope}
\end{figure*}

\paragraph{Model configurations.} \looseness=-1In this experiment, we train decoder-only Transformer language models with different positional encoding techniques while keeping all the other configurations the same. For Sinusoidal PE, we follow \citet{vaswani2017attention} to set the hyperparameters in sine and cosine functions. For RoPE and XPOS, we follow \citet{rope,sun2022length} to set the hyperparameters in the rotary matrix respectively. For Randomized RoPE, we set the extended positions $4$ times of the training length. We also conducted experiments on the extended positions $16$ times of the training length in Figure~\ref{fig:app_randomized_rope}, which shows performance degradation. For ALiBi, we follow \citet{alibi} to set the slope values in each attention head. For the intra segment encoding of our BiPE, we use the learnable absolute positional encoding. For the inter segment encoding of our BiPE-RoPE, the hyperparameters are kept the same as~\cite{rope}. For the inter segment encoding of our BiPE-ALiBi, the slope values are set to 96 times of the original ALiBi's setting. Other model configurations are provided in Table~\ref{tab:app_lm_model}.

\paragraph{Training recipes.} The next token prediction objective~\cite{brown2020language} is adopted for language model training. All models are trained on the Pile dataset\footnote{We use a copy of the Pile dataset with all copyrighted contents removed: \url{https://huggingface.co/datasets/monology/pile-uncopyrighted}.}~\cite{gao2020pile} with a total sequence length of 1024. The training recipes are shown in Table~\ref{tab:app_lm_train}. All models are trained on 8 NVIDIA A100 GPUs.

\subsection{Integrating BiPE with Fine-tuning Strategies}\label{app_exp:yarn}

\begin{table}[!ht]
\centering
\begin{minipage}{.5\textwidth}
\caption{Finetuning recipes for the YaRN strategy.}\label{tab:app_yarn_finetune}
\centering
\begin{tabular}{ccc}
\toprule
Batch size & & $64$ \\
Total training Steps & & $500$  \\
Dropout & & $0.0$  \\
Weight decay & & $0.01$  \\
Optimizer & & AdamW \\
Learning rate & & $2\mathrm{e}-5$   \\
\bottomrule
\end{tabular}
\end{minipage}%
\hfill
\begin{minipage}{.5\textwidth}
\caption{Finetuning recipes for long context benchmark.}\label{app:exp_scrolls_finetune}
\centering
\begin{tabular}{ccc}
\toprule
Batch size & & $64$ \\
Total training steps & & $5000$  \\
Dropout & & $0.0$  \\
Weight decay & & $0.01$  \\
Optimizer & & AdamW \\
Learning rate & & $1\mathrm{e}-5$   \\
\bottomrule
\end{tabular}
\end{minipage}
\end{table}

\paragraph{Model configurations.} In this experiment, we use the YaRN strategy to fine-tune pre-trained language models with RoPE and our BiPE-RoPE. All the model configurations are the same as those in Table~\ref{tab:app_lm_model}.

\paragraph{Fine-tuning recipes.} We set the scale factor in YaRN to 16 and fine-tune models using the next token prediction task for 500 steps on the Pile~\cite{gao2020pile} dataset with a sequence length of 4096. The finetuning recipes are shown in Table~\ref{tab:app_yarn_finetune}. All models are fine-tuned on 8 NVIDIA A100 GPUs.

\subsection{Long Context Benchmark}\label{app_exp:long_context_benchmark}

\paragraph{Model configurations.} In this experiment, we fine-tune pretrained language models with different positional encoding methods on SCROLLS~\cite{shaham2022scrolls}. It is a long context benchmark that consists of seven distinct datasets covering different tasks, e.g, Question-Answering (Qasper~\cite{dasigi2021qasper}, NarrativeQA~\cite{kocisky2018narrativeqa}, and QuALITY~\cite{pang-etal-2022-quality}), Natural Language Inference (ContractNLI~\cite{koreeda-manning-2021-contractnli-dataset}) and Summarization (QMSum~\cite{zhong2021qmsum}, SummScreenFD~\cite{chen-etal-2022-summscreen}, and GovReport~\cite{huang2021govreport}). All the model configurations are the same as those in Table~\ref{tab:app_lm_model}.

\paragraph{Fine-tuning recipes.} We fine-tune models using the next token prediction objective on each task with a sequence length of 8192. The finetuning recipes are provided in Table~\ref{app:exp_scrolls_finetune}. The model checkpoint that achieves the best performance on the validation set is selected for the final evaluation. The test results are obtained from the official SCROLLS website\footnote{\url{https://www.scrolls-benchmark.com/submission}}. All models are fine-tuned on 8 NVIDIA A100 GPUs.

\end{document}